\documentclass[10pt,twocolumn,letterpaper]{article}

\usepackage{cvpr}              %
\usepackage{booktabs}
\usepackage{multirow}
\usepackage{adjustbox}
\usepackage{xcolor}
\usepackage{rotating}
\usepackage[table]{xcolor}
\usepackage{pifont}
\newcommand{\cmark}{\ding{51}}
\newcommand{\xmark}{\ding{55}}
\usepackage{tikz}
\usepackage{adjustbox}
\usepackage{colortbl}
\usepackage[utf8]{inputenc}
\usepackage{enumitem}
\usepackage{xcolor}
\usepackage{mdframed}
\usepackage{pgfplots}

\usepackage[most]{tcolorbox}

\definecolor{cvprblue}{rgb}{0.21,0.49,0.74}
\usepackage[pagebackref,breaklinks,colorlinks,allcolors=cvprblue]{hyperref}

\def\paperID{xxxxx} %
\def\confName{CVPR}
\def\confYear{2026}

\title{LottieGPT: Tokenizing Vector Animation for Autoregressive Generation}

\usepackage{listings}
\usepackage{xcolor}

\lstdefinestyle{json}{
    basicstyle=\tiny\ttfamily,
    numbers=left,
    numberstyle=\tiny\color{gray},
    stepnumber=1,
    numbersep=5pt,
    backgroundcolor=\color{white},
    showspaces=false,
    showstringspaces=false,
    showtabs=false,
    frame=single,
    tabsize=2,
    breaklines=true,           %
    breakatwhitespace=false,
    escapeinside={(*@}{@*)},
    stringstyle=\color{blue},
    keywordstyle=\color{purple},
    commentstyle=\color{green}
}

\usepackage{listings}
\usepackage{xcolor}

\lstdefinestyle{json}{
    basicstyle=\tiny\ttfamily,
    breaklines=true,
    breakatwhitespace=false,
    columns=flexible,
    keepspaces=true,
    showstringspaces=false,
    frame=single,
    numbers=left,
    numberstyle=\tiny\color{gray},
    commentstyle=\color{green!50!black},
    stringstyle=\color{blue},
    keywordstyle=\color{purple},
    literate=
        *{:}{{{\color{red}{:}}}}{1}
        {,}{{{\color{red}{,}}}}{1}
        {\{}{{{\color{blue}{\{}}}}{1}
        {\}}{{{\color{blue}{\}}}}}{1}
        {[}{{{\color{blue}{[}}}}{1}
        {]}{{{\color{blue}{]}}}}{1},
}

\author{
Junhao Chen\textsuperscript{1,2 *} \quad
Kejun Gao\textsuperscript{2 *} \quad
Yuehan Cui\textsuperscript{2} \quad
Mingze Sun\textsuperscript{1} \quad
Mingjin Chen\textsuperscript{4}\\
Shaohui Wang\textsuperscript{1} \quad
Xiaoxiao Long\textsuperscript{5} \quad
Fei Ma\textsuperscript{6} \quad
Qi Tian\textsuperscript{6} \quad
Ruqi Huang\textsuperscript{1 \textdagger}\quad
Hao Zhao\textsuperscript{2,3 \textdagger} \\
\\
\textsuperscript{1}Shenzhen International Graduate School, Tsinghua University \quad 
\textsuperscript{2}AIR, Tsinghua University \\
\textsuperscript{3}BAAI \quad
\textsuperscript{4}The Hong Kong Polytechnic University \quad
\textsuperscript{5}Nanjing University \quad \\
\textsuperscript{6}Guangdong Laboratory of Artificial Intelligence and Digital Economy (SZ)
}

\begin{document}

\twocolumn[{
\renewcommand\twocolumn[1][]{#1}
\maketitle
\begin{center}
    \vspace{-7mm}
    Project Page: \url{https://lottiegpt.github.io/} \\
    \vspace{-6mm}
\end{center}
\
\begin{center}
    \captionsetup{type=figure}
    \centerline{\includegraphics[width=\linewidth]{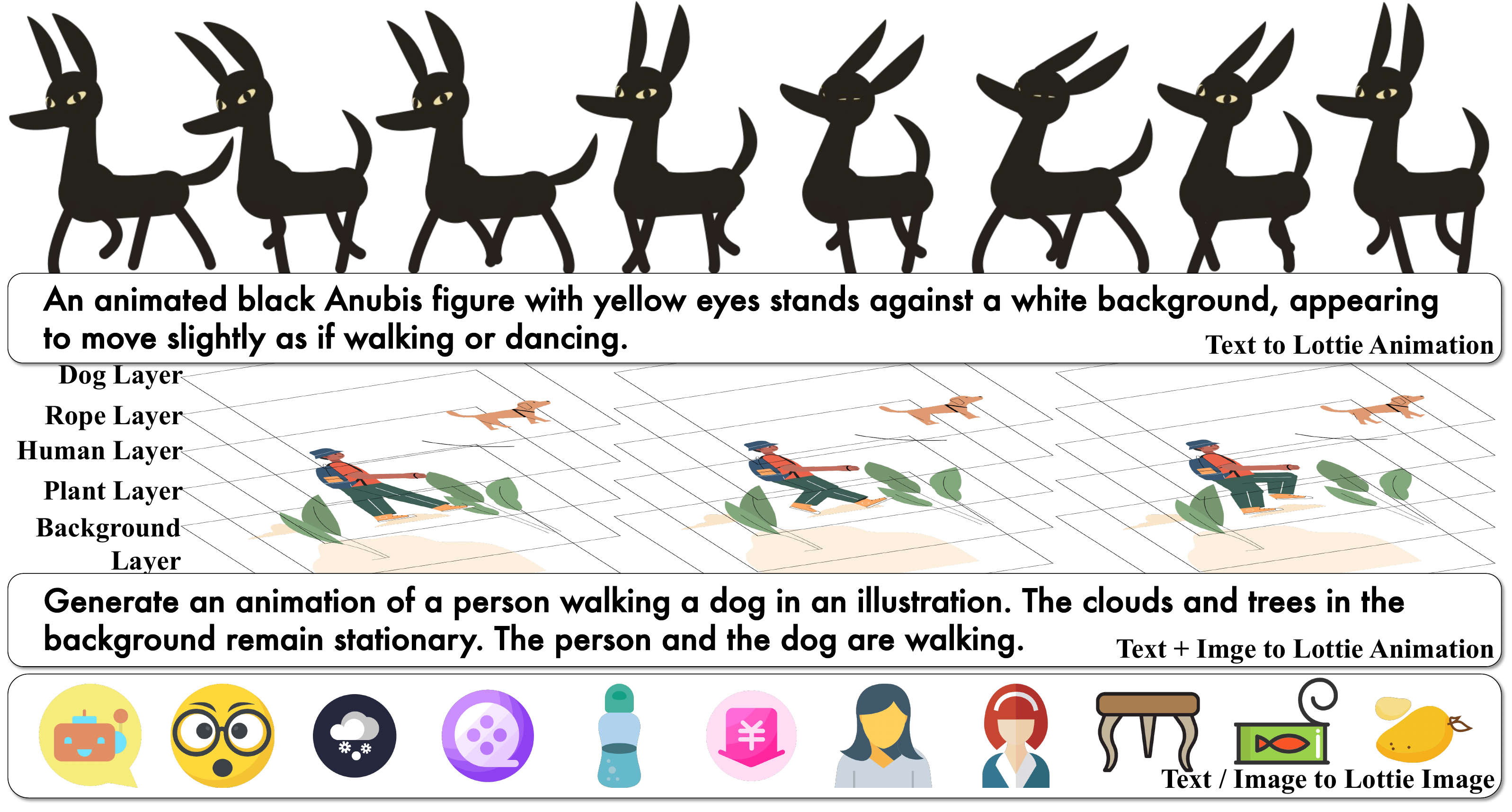}}
    \caption{
    \textbf{LottieGPT generates editable vector animations from diverse inputs.} 
    Unlike existing models that produce fixed-resolution raster videos, we generate resolution-independent vector graphics and animations from text, image, or keyframes.
    Our outputs \textbf{scale infinitely}, and \textbf{enable direct editing of shapes and motion}. These capabilities are impossible with pixel-based methods.
    }
    \label{fig:teaser}
    \vspace{-2mm}
\end{center}
}]

\renewcommand{\thefootnote}{} 
\footnotetext{
    \textsuperscript{*} Equal Contribution. \textsuperscript{\textdagger} Corresponding Author.
}

\setcounter{footnote}{0}
\renewcommand{\thefootnote}{\arabic{footnote}}

\begin{abstract}

Despite rapid progress in video generation, existing models are incapable of producing vector animation, a dominant and highly expressive form of multimedia on the Internet. Vector animations offer resolution-independence, compactness, semantic structure, and editable parametric motion representations, yet current generative models operate exclusively in raster space and thus cannot synthesize them. Meanwhile, recent advances in large multimodal models demonstrate strong capabilities in generating structured data such as slides \cite{ge2025autopresent}, 3D meshes \cite{tang2024edgerunner}, LEGO sequences \cite{pun2025generating}, and indoor layouts \cite{mao2025spatiallm}, suggesting that native vector animation generation may be achievable. In this work, we present the first framework for tokenizing and autoregressively generating vector animations. We adopt Lottie, a widely deployed JSON-based animation standard, and design a tailored Lottie Tokenizer that encodes layered geometric primitives, transforms, and keyframe-based motion into a compact and semantically aligned token sequence. To support large-scale training, we also construct \textbf{LottieAnimation-660K}, the largest and most diverse vector animation dataset to date, consisting of 660k real-world Lottie animation and 15M static Lottie image files curated from broad Internet sources. Building upon these components, we finetune Qwen-VL to create \textbf{LottieGPT}, a native multimodal model capable of generating coherent, editable vector animations directly from natural language or visual prompts. Experiments show that our tokenizer dramatically reduces sequence length while preserving structural fidelity, enabling effective autoregressive learning of dynamic vector content. LottieGPT exhibits strong generalization across diverse animation styles and outperforms previous state-of-the-art models on SVG generation (a special case of single-frame vector animation). %

\end{abstract}
    
\section{Introduction}
\label{sec:intro}

Recent breakthroughs in text-to-video generation have advanced the fidelity, coherence, and controllability of pixel space synthesis. Models such as Sora~\cite{openai2024sora2} and Kling~\cite{kling} now produce photorealistic footage with compelling dynamic consistency. Yet despite this rapid progress, \textbf{none} of these systems can generate vector animation, a dominant medium underlying modern digital communication, UI/UX motion design, educational content, product illustration, branding, and countless web applications. Vector animations offer properties fundamentally absent from raster video: infinite resolution, structural editability, layered organization, parametric motion, compact file size, and semantic manipulability. These characteristics make vector animation central to professional workflows, from After Effects motion graphics to Lottie-based mobile interfaces.

Meanwhile, vision language models (VLMs) have shown surprising competence in generating structured representations. Models can now autogenerate slide decks \cite{ge2025autopresent}, 3D meshes \cite{tang2024edgerunner}, garment pattern~\cite{weng2026garmentgpt}, LEGO assembly sequences \cite{pun2025generating}, indoor layouts \cite{mao2025spatiallm}, and other discrete or hierarchical content. These advances reveal an important \textbf{trend}: VLMs are increasingly capable of manipulating symbolic structures instead of only pixel grids. Since vector animations are themselves structured programs (composed of hierarchical layers, geometric primitives, keyframes, and easing functions), this capability suggests that native vector animation generation could be within reach.

However, achieving this requires solving two fundamental challenges. The \textbf{first} and most critical is the tokenizer: converting rich, temporally organized vector animation into a token sequence suitable for autoregressive modeling. Unlike static SVGs or meshes, vector animations contain both hierarchical structure and time-dependent transformation logic. To address this, we adopt the widely deployed Lottie format, a JSON-based representation used at scale across web. Lottie encodes animations as layered shapes with parametric transforms, keyframe schedules, interpolators, and easing curves. We design the first Lottie Tokenizer, which decomposes a Lottie file into a compact, semantically aligned set of tokens that capture geometric primitives, hierarchical grouping, animated property curves, and interpolation settings. Our tokenizer stores keyframes and easing functions rather than dense per-frame data, dramatically reducing sequence length while preserving structural fidelity. No previous tokenizer in structured data generation has attempted to unify \textbf{both hierarchical geometry and temporal motion} in a single token stream.

The \textbf{second} major challenge is data. Before our work, no large-scale vector animation dataset existed due to the dominance of rasterized video on the web. We build the first Lottie animation corpus, Lottie-660K, consisting of 660K high-quality, diverse animations exported from After Effects via the Bodymovin pipeline, along with extensive cleaning, standardization, and JSON simplification. In addition, we curate 15M static vector graphics converted into Lottie format, enabling a progressive static-to-dynamic training strategy. Together, these datasets constitute the largest resource ever built for vector animation research.

With these components, we finetune Qwen2.5-VL to create \textbf{LottieGPT}, the first multimodal model capable of generating fully editable vector animations from text, images, or keyframes. LottieGPT demonstrates robust performance across diverse in-the-wild scenarios: icon dynamics, UI transitions, illustrative cartoons, and multi-layered scene animations. When evaluated on SVG generation (a special case of single-frame vector animation), our model achieves new state-of-the-art performance, confirming that temporal modeling strengthens even static vector understanding.

Our contributions are four-fold:
\begin{enumerate}
    \item We present the first framework for native vector animation generation, moving beyond video to resolution-independent, semantically editable motion graphics.
    \item We design the first tokenizer capable of encoding hierarchical geometric primitives, transforms, keyframes, and temporal dynamics into a compact token sequence.
    \item We construct 660K Lottie animation and 15M Lottie static graphics, establishing the largest and most diverse resource for vector animation learning, and propose LottieBench, the first benchmark for vector animation.
    \item We develop and train the first multimodal model (LottieGPT) for autoregressive vector animation generation, demonstrating strong in-the-wild performance and achieving new SOTA on SVG generation.
\end{enumerate}

\begin{figure*}[t]
    \centering
    \includegraphics[width=1\linewidth]{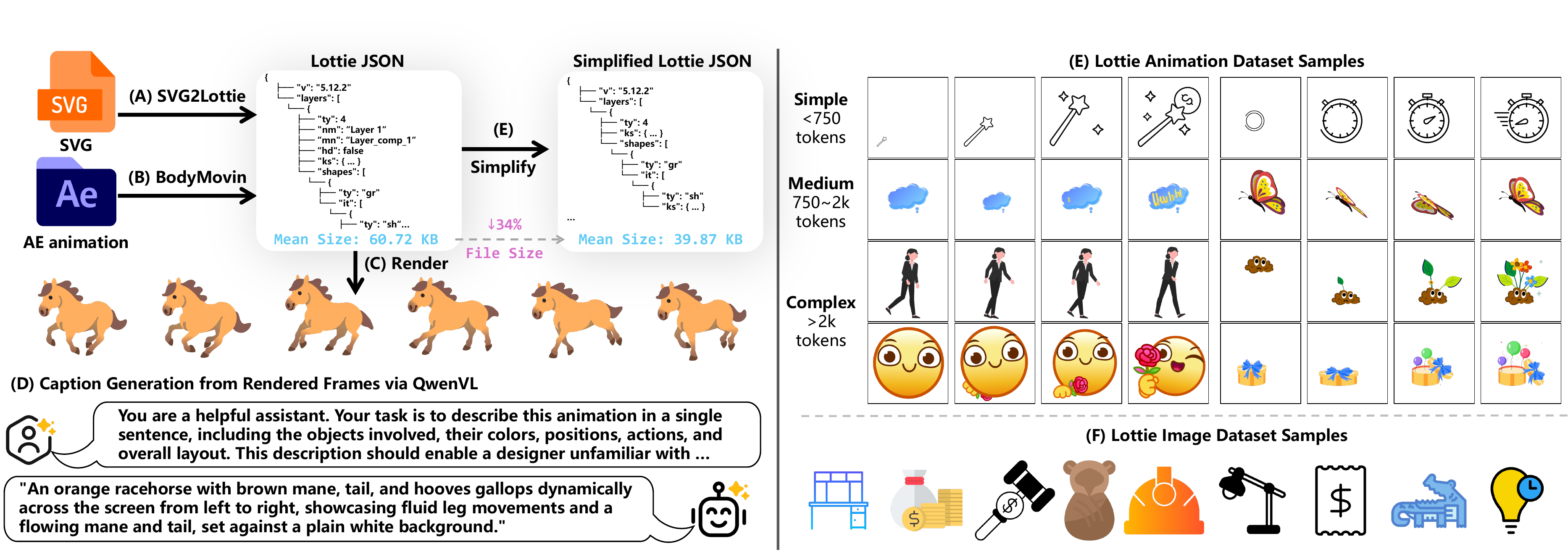}
    \caption{\textbf{Data curation pipeline.}
    We collected 10M SVG resources and 660K After Effects (AE) animation resources from the internet, then converted them to Lottie Json format, filtered them using simplification algorithms that do not affect rendering results, and used QwenVL to generate text labels for vector graphics and vector animations.
    }
    \label{fig:datacurationpipe}
    \vspace{-2mm}
\end{figure*}

\section{Related Works}
\subsection{Pixels based Image and Video Generation}

Diffusion models have revolutionized visual content generation, achieving unprecedented capabilities in image synthesis~\cite{wu2025qwenimagetechnicalreport,flux1,StableDiffusion,zhang2024transparent,zhang2023adding,xing2024svgfusion} and video synthesis~\cite{videoworldsimulators2024,kling,hunyuanvideo,google2024veo31,guo2023animatediff,chen2025dancetogether,xu2024easyanimatehighperformancelongvideo,yang2024cogvideox,hong2022cogvideo,lin2025interanimate,chen2026hvg3dbridgingrealsimulation}. 
Despite their impressive visual quality, these pixel-based approaches produce non-editable fixed-resolution outputs that require substantial storage, cannot scale without quality degradation, and lack semantic editability. 
While recent editing methods~\cite{wu2025qwenimagetechnicalreport,ceylan2023pix2video,vace,liu2024video,chen2023soulstyler,brooks2023instructpix2pix,zhang2023adding} have partially addressed these issues, the raster nature fundamentally constrains their applicability in professional design workflows requiring iterative refinement and precise control.
Our work addresses these limitations by shifting the generation paradigm from pixel space to structured vector animation space.

\subsection{Structured Data Generation}

Recent research has explored generating structured representations that offer interpretability, compactness, and editability advantages over pixel-based outputs. 
In 3D generation, methods like LLaMA-Mesh~\cite{wang2024llama}, MeshGPT~\cite{siddiqui2024meshgpt}, and EdgeRunner~\cite{tang2024edgerunner} generate mesh structures as tokenized sequences~\cite{gao2025meshart,hao2024meshtron,lionar2025treemeshgptartisticmeshgeneration} through autoregressive modeling. 
Existing 3D animation production paradigms first generate 3D meshes~\cite{xiang2025trellis2,ye2025hi3dgen,lei2025hunyuan3d,chen2025idea23d,hong2023lrm,miao2026framessequencestemporallyconsistent,tang2024lgm,xiang2025structured,long2024wonder3d,chen2024ultraman,qiu2025lhm}, then create skeletons and animations~\cite{sun2025drive,song2025magicarticulate,jiang2023motiongpt,dai2024motionlcm,xiao2025motionstreamer,fan2025go,Chen_2025_ICCV,han2025atom,guo2025make,song2025puppeteer}, essentially represent a form of 3D vector animation. We adapt this 3D animation production paradigm to 2D animation generation.
Domain-specific approaches include DeepCAD~\cite{wu2021deepcad} for parametric CAD design~\cite{khan2024text2cad,alrashedygenerating,liu2025hola} and BrickGPT~\cite{pun2025generating} for physically feasible LEGO structures. In 2D vector graphics~\cite{jain2023vectorfusion,xing2024svgdreamer}, StarVector~\cite{xing2025star} and OmniSVG~\cite{yang2025omnisvg} generate SVG code by treating it as a code synthesis task.
YOLaT~\cite{dou2024hierarchically} performs scientific image understanding at the SVG level.
Some models generate structured text captions from video or 3D inputs~\cite{bai2025qwen2,ye2024mmad,dinghint,li2022toist,xue2023ulip,chiimpromptu,wang2025n3d,jin2024tod3cap}.
However, these methods remain confined to static outputs, lacking temporal modeling capabilities essential for animations. These capabilities include coherent motion, keyframe coordination, and temporal property parameterization.

\subsection{Vector Graphics and Animation Generation}

Vector graphics generation~\cite{wu2023iconshop,zhang2024text,chen2025svgbuilder,wu2025layerpeeler,reddy2021im2vec,zhao2024vector} has evolved along two trajectories: optimization-based methods~\cite{li2020differentiable,jain2023vectorfusion} suffer from lengthy computation, while VLM-based methods~\cite{xing2025star,yang2025omnisvg,wu2025chat2svg,xing2025empowering} achieve better editability by generating SVG code directly. However, both remain confined to static outputs.
Vector animation generation faces three challenges: limited format expressiveness, data scarcity, and temporal coordination complexity. Existing methods rely on simple interpolation~\cite{carlier2020deepsvg,LIVESKETCH,gao2025linr,tseng2024keyframer,mateja2023animatesvg,dalstein2015vector} or domain-specific solutions~\cite{wu2025aniclipart,liu2024dynamic,iluz2023word,ma2025mover,zhang2023editing}, failing to generalize.
We address these limitations using the Lottie format, which enables hierarchical composition and parametric control. By formulating generation as structured code synthesis rather than pixel-space video generation, we achieve lightweight (10-50× smaller), resolution-independent, and fully editable animations.

\begin{figure*}
    \centering
    \includegraphics[width=1\linewidth]{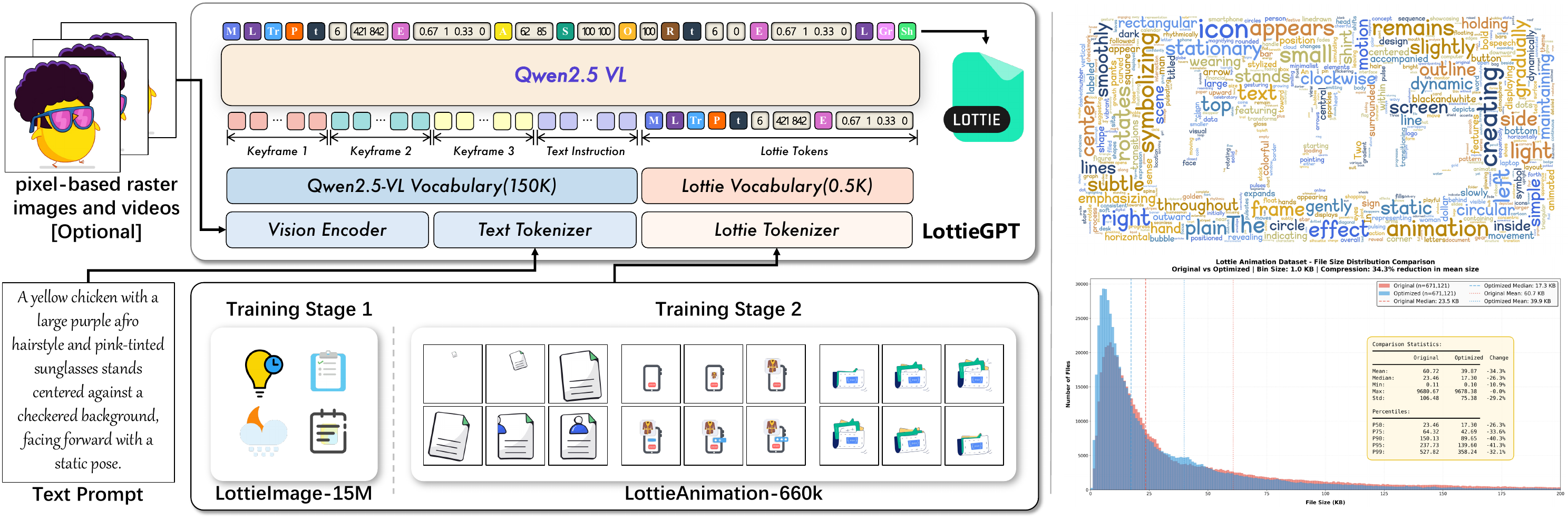}
    \caption{\textbf{Overview of LottieGPT.} LottieGPT is built upon the pre-trained vision-language model Qwen2.5-VL and incorporates a Lottie tokenizer. The model encodes both text and image inputs as prefix tokens, while the Lottie tokenizer encodes vector animation commands into a unified representation space. We first train the model on static Lottie images, followed by training on Lottie animations.}
    \label{fig:pipeline}
    \vspace{-2mm}
\end{figure*}

\section{Data Curation Pipeline}

While pixel-based visual data and static vector graphics datasets (e.g., SVG) are abundant, large-scale vector animation datasets remain scarce. Due to SVG's limited support for complex timeline-based animations, we turn to the After Effects (AE) ecosystem, leveraging the Bodymovin plugin\footnote{\url{https://aescripts.com/bodymovin/}} to export AE animations into Lottie JSON format, which is a lightweight, cross-platform representation supporting keyframes, path morphing, and color gradients.

We collect AE source files from public resources and convert them via an automated pipeline. We integrate 15M static vector graphics, uniformly converted to Lottie representation, to support a progressive ``\textbf{static-first, then dynamic}'' training strategy. For multimodal generation, we use BLIP-2~\cite{li2023blip} for static graphics captions and Qwen2.5-VL 32B~\cite{bai2025qwen2} for temporally aligned animation descriptions after rendering to video.

We apply rigorous filtering: removing rendering failures and visual anomalies (blank frames, misalignment, flickering), and simplifying Lottie JSON by removing redundant fields (comments, unused properties, debug metadata) while preserving rendering fidelity. This reduces sequence length by ~34\% without affecting quality. 
Tab.~\ref{tab:dataset_comparison} compares our datasets with existing vector graphics datasets.
Fig.~\ref{fig:datacurationpipe} illustrates our dataset construction pipeline and sample examples.
\textbf{See Appendix~\ref{supp_lottiedataset} for the details.}

\begin{table}[t]
\centering
\caption{Comparison of vector graphics and animation datasets. 
}
\label{tab:dataset_comparison}
\resizebox{\columnwidth}{!}{
\begin{tabular}{l|c|c|c|c}
\hline
\textbf{Dataset} & \textbf{Format} & \textbf{Type} & \textbf{Size} & \textbf{Captions} \\
\hline
\hline
\multicolumn{5}{c}{\textit{Static Vector Graphics Datasets}} \\
\hline
SVG-Stack~\cite{rodriguez2025starvector} & SVG & Icons/Graphics & 2.3M & \cmark \\
SVGX~\cite{xing2024llm4svg,xing2024svgfusion} & SVG & Icons/Illustrations & 1M & \cmark \\
OmniSVG~\cite{yang2025omnisvg} & SVG & Icons/Illustrations & 2M & \cmark \\
SVGFind~\footnote{https://huggingface.co/datasets/nyuuzyou/svgfind} & SVG & Icons/Graphics & 4M & \xmark \\
DeepSVG Icons8~\cite{carlier2020deepsvg} & SVG & Icons & 100K & \cmark \\
\hline
\multicolumn{5}{c}{\textit{Our Proposed Dataset}} \\
\hline
\textbf{LottieSVG} & SVG/Lottie & Icons/Illustrations & \textbf{10M} & \cmark \\
\textbf{LottieAnimation} & Lottie & Animations & \textbf{660K} & \cmark \\
\textbf{LottieImage} & Lottie & Icons/Illustrations & \textbf{15M} & \cmark \\
\hline
\end{tabular}
}
\vspace{-4mm}
\end{table}

\section{Methods}
\subsection{Overview}

Our LottieGPT framework builds upon the Qwen2.5-VL architecture~\cite{bai2025qwen2}, a vision-language model that excels in processing both visual and textual inputs. To enable native vector animation generation, we extend the model vocabulary with specialized Lottie tokens and design a compact animation tokenization scheme. As illustrated in Fig.~\ref{fig:pipeline}, our approach consists of three key components: (1) a Lottie Tokenizer that converts JSON-based animations into discrete token sequences, (2) a vision-language backbone that processes multimodal inputs, and (3) a two-stage training strategy that progressively learns from static graphics to dynamic animations.
Unlike existing approaches that generate raster videos or frame-by-frame SVG sequences, LottieGPT directly generates structured vector animations in Lottie format. This enables resolution-independent, compact, and fully editable outputs. Our model takes text descriptions, reference images, or keyframe videos as input and autoregressively generates Lottie token sequences that can be decoded into fully functional vector animations.

\begin{figure*}
    \centering
    \includegraphics[width=1\linewidth]{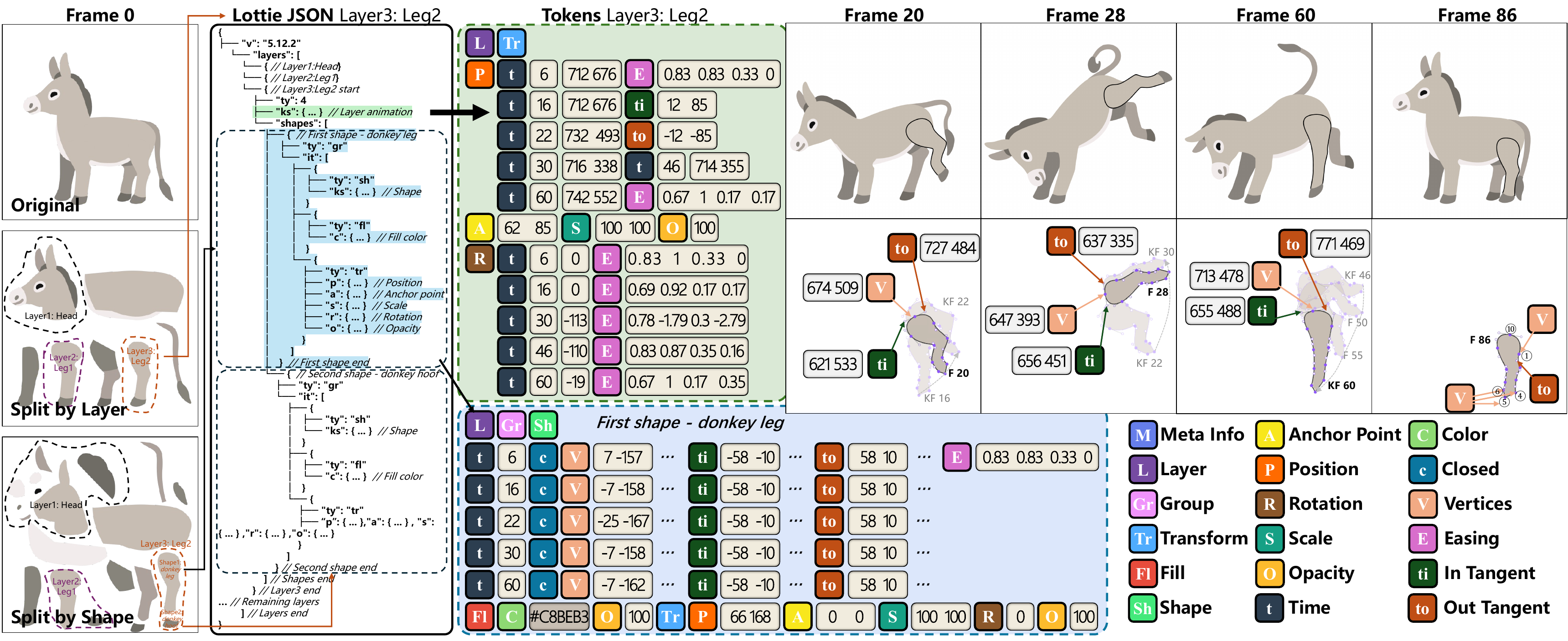}
    \caption{Unlike raster pixel-based videos or frame-by-frame saved SVGs, the Lottie Tokenizer only stores \textbf{keyframes} and \textbf{interpolation} methods, which significantly reduces the number of tokens required to represent an animation. In the figure, \textbf{\textit{KF}} denotes keyframes, while \textit{\textbf{F}} represents frames obtained through easing-based animation interpolation.}
    \label{fig:lottie_tokenizer}
    \vspace{-2mm}
\end{figure*}

\subsection{Compact Animation Tokenization}

Autoregressive models process information as discrete token sequences. Therefore, compact tokenization is crucial, as it enables accurate representation of information with fewer tokens. Unlike text tokenizers that achieve highly compact and lossless compression by combining subword units into single tokens, tokenization techniques used in prior autoregressive vector graphics generation work suffer from two main issues. First, decomposing SVGs into basic atomic commands~\cite{yang2025omnisvg,carlier2020deepsvg} loses semantic information at the layer group level. Second, existing vector graphics tokenizers cannot represent vector animations, as these methods operate solely on static vector frame images.

Therefore, we design a specialized tokenizer for Lottie animations that hierarchically encodes layers, shapes, and temporal keyframes. Unlike previous SVG tokenizers that treat code as plain text, our tokenizer exploits the structured nature of Lottie JSON to achieve superior compression while preserving the semantic information inherent in the original layers.

\subsubsection{Hierarchical Structure Encoding}

A Lottie animation follows a strict hierarchical structure: \textbf{Animation Meta} $\rightarrow$ \textbf{Assets} $\rightarrow$ \textbf{Layers} $\rightarrow$ \textbf{Shapes} $\rightarrow$ \textbf{Properties}. We encode this hierarchy using special tokens that mark boundaries and relationships:

\noindent\textbf{Animation Meta Encoding.} We first encode global animation metadata using a compact representation:
\begin{verbatim}
<|M|><|v|>"5.9.5"<|fr|>30<|ip|>0<|op|>90
    <|w|>512<|h|>512<|ddd|>0
\end{verbatim}
This single-line encoding captures version, frame rate, in-point, out-point, dimensions, and 3D flag. This information would require multiple lines in plain JSON text tokenization.

\noindent\textbf{Layer Encoding.} Each layer is encoded with its type, transform, and child elements:
\begin{verbatim}
<|LAYER|><|ty|>4<|ip|>0<|op|>90<|st|>0
    <|bm|>0<|LAYER_KS|>...
\end{verbatim}
Unlike naive text tokenization, we use specialized tokens (\texttt{<|LAYER|>}, \texttt{<|ty|>}) that directly correspond to Lottie schema, enabling the model to learn structural patterns rather than arbitrary text sequences.

\noindent\textbf{Shape Encoding.} Within each layer, we encode geometric shapes using a type-specific format. For example, a path shape:
\begin{verbatim}
<|ITEM_sh|><|KS_STATIC|><|i|>0 0 -10 5
    <|o|>0 0 10 -5<|v|>100 50 150 75<|c|>
\end{verbatim}
This representation encodes in-tangents \texttt{i}, out-tangents \texttt{o}, vertices \texttt{v}, and closed flag \texttt{c} in vertex order, following a linearized encoding scheme.

Beyond basic shape paths, our Lottie Tokenizer directly encodes complex Lottie shape primitives including Ellipse, Fill, Gradient, Gradient Stroke, Group, PolyStar, Rectangle, Rounded Corners, and Stroke, without decomposing them into independent line segments as in OmniSVG~\cite{yang2025omnisvg}.

\subsubsection{Keyframe-Based Motion Compression}
The key innovation distinguishing our tokenizer from prior work is \textbf{keyframe-based temporal compression}. As illustrated in Fig.~\ref{fig:lottie_tokenizer}, instead of storing every frame like raster video or frame-by-frame SVGs, we encode only \textbf{keyframes} and \textbf{interpolation functions}.

\noindent\textbf{Unified Property Animation Encoding.}
Our tokenizer employs a unified encoding scheme for all animated properties. Each animated property follows the same structural pattern:
\begin{verbatim}
<|PROP_ANIMATED|><|PROP_KF_START|>
  <keyframe_1><keyframe_2>...<keyframe_n>
<|PROP_KF_END|>
\end{verbatim}

\noindent\textbf{Transform Animation.}
Consider a layer transform with animated position, rotation, and opacity. A typical Lottie transform contains five core properties: \textit{position} (p), \textit{anchor point} (a), \textit{scale} (s), \textit{rotation} (r), and \textit{opacity} (o). Each property can be either independently animated or a static value.

\noindent\textbf{Keyframe Structure.}
Each keyframe encodes three essential components:
\begin{itemize}[leftmargin=*,noitemsep,topsep=0pt]
    \item \texttt{<|t|>}: Time in frames that specifies when this keyframe occurs
    \item Value tokens: Property-specific encoding based on dimensionality
    \item \texttt{<|ease|>}: Cubic Bézier easing function (optional)
    \begin{itemize}[noitemsep]
        \item Empty tag indicates default ease-in-out curve
        \item Four parameters $(i_x, i_y, o_x, o_y)$ for custom curves
        \item Omitted for the final keyframe (no interpolation needed)
    \end{itemize}
\end{itemize}

\noindent\textbf{Compression Analysis.}
This representation achieves \textbf{dramatic compression}. For example, as shown in Fig.~\ref{fig:lottie_tokenizer}, consider a 100-frame animation with smooth transitions across 6 keyframes.
Moreover, the easing function \texttt{<|ease|>} encodes Bézier control points that define motion curves, which is \textbf{critical information for professional animation quality} that is entirely absent in raster representations. This allows the same keyframes to produce drastically different motion feels (linear, ease-in, bounce, etc.) without additional data.

\noindent\textbf{Advantages.}
Compared to frame-based approaches, our keyframe encoding offers: (1) \textbf{temporal scalability}, where compression ratio improves with duration (98\% for 300 frames/5 keyframes), (2) \textbf{motion quality preservation}, where easing curves are first-class primitives rather than approximations, (3) \textbf{editability}, where individual keyframes can be modified without sequence reconstruction, (4) \textbf{resolution independence}, where coordinate-based values scale freely to any resolution, and (5) \textbf{semantic integrity}, where shapes and layers are encoded as complete hierarchical units rather than decomposed into atomic primitives~\cite{yang2025omnisvg}, preserving compositional structure for VLM learning.

\noindent\textbf{Tokenization and Detokenization.}
We traverse the Lottie JSON tree depth-first, encoding metadata, assets, layers, and shapes with special tokens. Each property is encoded as either static (\texttt{<|\_STATIC|>}) or animated (\texttt{<|\_ANIMATED|>}) based on keyframe detection, with property-specific compression applied (colors$\rightarrow$hex, motion$\rightarrow$Bézier). Detokenization reconstructs the JSON via recursive parsing, achieving \textbf{lossless roundtrip}: decoded animations render identically to originals.

\subsection{Static-to-Dynamic Training}

Following curriculum learning in LLMs, we adopt a \textbf{two-stage training approach} that first teaches static vector graphics generation before introducing temporal dynamics, since Lottie keyframes share the same representation as static graphics.

\noindent\textbf{Stage 1: Static Vector Graphics.} We train on static Lottie images (converted from SVG) with 50\% text-only data (text-to-Lottie generation) and 50\% multimodal data (image-to-Lottie generation). This stage teaches fundamental vector composition: shapes, fills, strokes, transforms, and hierarchical structure.

\noindent\textbf{Stage 2: Vector Animation.} We introduce temporal dynamics using Lottie animations with 34\% text-only, 33\% text + first-frame image, and 33\% text + video (keyframes). This mixture enables flexible conditioning at inference, where users can provide text descriptions, reference images for style guidance, or video keyframes for motion transfer.

\noindent\textbf{Why Two Stages?} Early experiments with joint training on mixed static+dynamic data showed unstable convergence. 
This is because each sample in LottieAnimation contains significantly more tokens than those in LottieImage, making it difficult for the model to simultaneously learn spatial composition and temporal coordination.
By first mastering static graphics, the model develops strong priors for vector primitives and hierarchical organization, then focuses purely on motion modeling in Stage 2.

\noindent\textbf{Training Objective.} Standard causal language modeling with cross-entropy loss on next-token prediction. The model learns to generate valid Lottie token sequences by minimizing:
\begin{equation}
\mathcal{L} = -\sum_{i=1}^{N} \log P(t_i \mid t_{<i}, \mathbf{c})
\end{equation}
where $t_i$ is the $i$-th token, $t_{<i}$ represents all previous tokens, and $\mathbf{c}$ denotes multimodal conditioning (text + image/video features).

\section{LottieBench}

The field of vector graphics already has numerous evaluation methods~\cite{nishina2024svgeditbench,zou2024vgbench,chen2025svgenius}, but the vector animation domain still lacks evaluation approaches.
To conduct a more comprehensive evaluation of Lottie animations, inspired by web code evaluation methods~\cite{iwbench}, we evaluate the generated Lottie JSON across five dimensions: visual level, structured data level, semantic level, and rendering success rate.

\subsection{Evaluation Data}

For animation generation, we stratify the test set into three difficulty levels by token count (Fig.~\ref{fig:datacurationpipe}): Simple (150 samples), Medium (40 samples), and Complex (40 samples), totaling 230 unseen animations. For static graphics generation, we evaluate on 400 randomly selected unseen samples without difficulty stratification, as static graphics are consistently short, typically under 200 tokens.

\begin{table*}[htbp]
\centering
\caption{Lottie image and animation generation with Text-Only, Text+Image input.}
\label{tab:animation_comp}
\begin{adjustbox}{max width=\textwidth}
\begin{tabular}{llccccccc}
\toprule
\textbf{Method} & \textbf{Model/Approach} & \multicolumn{4}{c}{\textbf{Visual Layer}} & \textbf{Structural Layer} & \textbf{Semantic Layer} & \textbf{Valid Rate} \\
\cmidrule(lr){3-6}
& & \textbf{CLIP↑} & \textbf{SSIM↑} & \textbf{LPIPS↓} & \textbf{DINOv2↑} & \textbf{JSON↑} & \textbf{CLIP-T↑} & \textbf{(\%)} \\
\midrule
\rowcolor{orange!20}
    \multicolumn{9}{c}{\textit{\textbf{Text-Only Input to Static Vector Graphics}}} \\
\bottomrule
\toprule
\multirow{1}{*}{Text2SVG} 
    & OmniSVG-7B & 0.8321 & 0.5627 & 0.5123 & 0.7274 & N/A & 0.2676 & N/A \\
\midrule
\rowcolor{gray!20}
    \textbf{Ours (Text Only)} & \textbf{LottieGPT-7B-Stage1} & \textbf{0.9331} & \textbf{0.8103} & \textbf{0.1755} & \textbf{0.8572} & \textbf{0.8239} & \textbf{0.2893} & \textbf{98.25\%} \\
\bottomrule
\toprule
\rowcolor{orange!20}
    \multicolumn{9}{c}{\textit{\textbf{Text+Image Input to Static Vector Graphics}}} \\
\midrule
\multirow{2}{*}{Image2SVG} 
    & StarVector-8B & 0.7662 & 0.3851 & 0.4647 & 0.5286 & N/A & 0.2454 & N/A \\
    & OmniSVG-7B & 0.9002 & 0.7051 & 0.2507 & 0.8481 & N/A & 0.2772 & N/A \\
\midrule
\rowcolor{gray!20}
    \textbf{Ours (Text+Image)} & \textbf{LottieGPT-7B-Stage1} & \textbf{0.9201} & \textbf{0.8151} & \textbf{0.1941} & \textbf{0.8514} & \textbf{0.8472} & \textbf{0.2871} & \textbf{97.50\%} \\
\bottomrule
\toprule
\rowcolor{orange!20}
    \multicolumn{9}{c}{\textit{\textbf{Text-Only Input to Vector Animation}}} \\
\midrule
\multirow{5}{*}{\shortstack[l]{Few-shot \\ (3-shot)}} 
& GPT-5 & 0.7682 & 0.8232 & 0.2528 & 0.5929 & 0.0358 & 0.2640 & 10.87\% \\
& Claude Sonnet 4.5 & 0.8485 & 0.8493 & 0.2496 & 0.7278 & 0.0522 & 0.3115 & 45.22\% \\
& Gemini 2.5 Pro & 0.7157 & 0.6574 & 0.2113 & 0.6400 & 0.0403 & 0.2565 & 30.00\% \\
& DeepSeek-V3.1 & 0.5425 & 0.6390 & 0.1358 & 0.3721 & 0.0399 & 0.1632 & 30.43\% \\
& Qwen3-235B & 0.8122 & 0.7973 & 0.3084 & 0.6637 & 0.0151 & 0.3185 & 28.26\% \\
\midrule
\multirow{1}{*}{\shortstack[l]{Finetuned w. Lottie Json}} 
& Qwen2.5-VL 7B  & 0.7456 & 0.7156 & 0.1714 & 0.4789 & 0.0089 & 0.2236 & 22.61\% \\
\midrule
\rowcolor{gray!5}
\textbf{Ours (Text Only)} & \textbf{LottieGPT-7B-Stage1} & \textbf{0.8156} & \textbf{0.8524} & \textbf{0.1925} & \textbf{0.7715} & \textbf{0.2347} & \textbf{0.2756} & \textbf{78.35\%} \\
\rowcolor{gray!20}
\textbf{Ours (Text Only)} & \textbf{LottieGPT-7B-Stage2} & \textbf{0.9566} & \textbf{0.9776} & \textbf{0.0366} & \textbf{0.9301} & \textbf{0.8062} & \textbf{0.2985} & \textbf{96.96\%} \\
\bottomrule
\toprule
\rowcolor{orange!20}
    \multicolumn{9}{c}{\textit{\textbf{Text + Image Input to Vector Animation}}} \\
\midrule
\multirow{3}{*}{\shortstack[l]{Few-shot \\ (3-shot)}} 
& Claude Sonnet 4.5 & 0.8499 & 0.9229 & 0.1273 & 0.7349 & 0.0043 & 0.3416 & 51.74\% \\
& Gemini 2.5 Pro & 0.8476 & 0.8311 & 0.2136 & 0.6980 & 0.0037 & 0.2862 & 32.17\% \\
& Qwen3-235B & 0.8162 & 0.9251 & 0.1721 & 0.7084 & 0.0028 & 0.2831 & 24.35\% \\
\midrule
\multirow{3}{*}{Image2Video} 
& Sora2 & 0.8903 & 0.8661 & 0.3004 & 0.8182 & N/A & N/A & N/A \\
& Kling & 0.8661 & 0.8542 & 0.2783 & 0.6261 & N/A & N/A & N/A \\
& Veo3.1 & 0.8487 & 0.5558 & 0.4417 & 0.6658 & N/A & N/A & N/A \\
\midrule
\rowcolor{gray!20}
\textbf{Ours (Text+Image)} & \textbf{LottieGPT-7B-Stage2} & \textbf{0.9743} & \textbf{0.9886} & \textbf{0.0278} & \textbf{0.9472} & \textbf{0.9195} & \textbf{0.3029} & \textbf{97.83\%} \\
\bottomrule
\end{tabular}
\end{adjustbox}
\vspace{-4mm}
\end{table*}

\subsection{Evaluation Metrics}

We evaluate generated Lottie animations across three complementary dimensions: \textbf{visual fidelity}, \textbf{structural correctness}, and \textbf{semantic alignment}.

\noindent\textbf{Visual-Level Metrics.}
We assess perceptual quality using standard image and video metrics: LPIPS~\cite{zhang2018unreasonable}, SSIM~\cite{wang2004image}, CLIP~\cite{radford2021learning}, and DINO~\cite{caron2021emerging}.

\noindent\textbf{Structural-Level Metrics.}
For methods that output Lottie JSON, we evaluate JSON structure consistency.
To evaluate structural correctness of generated Lottie JSON, we perform three steps:

\noindent\textit{Step 1: Flatten JSON hierarchy.} 
Traverse the nested JSON depth-first to extract all key-value pairs:
\begin{equation}
\Phi(\mathcal{J}) = \{(k, v) \mid k \in \mathcal{K}(\mathcal{J})\}
\end{equation}
where $k$ is a hierarchical path like \texttt{$layers[0].shapes[1].ty$}.

\noindent\textit{Step 2: Compare key sets.} 
Partition keys into common ($\mathcal{K}^c$), missing ($\mathcal{K}^m$), and extra ($\mathcal{K}^e$):
\begin{equation}
\mathcal{K}^c = \mathcal{K}^{\text{gt}} \cap \mathcal{K}^{\text{pred}}, \quad
\mathcal{K}^m = \mathcal{K}^{\text{gt}} \setminus \mathcal{K}^{\text{pred}}, \quad
\mathcal{K}^e = \mathcal{K}^{\text{pred}} \setminus \mathcal{K}^{\text{gt}}
\end{equation}
Compute Key-$F_1$ to measure topology correctness:
\begin{equation}
\text{Key-}F_1 = \frac{2|\mathcal{K}^c|}{|\mathcal{K}^{\text{gt}}| + |\mathcal{K}^{\text{pred}}|}
\end{equation}

\noindent\textit{Step 3: Measure value consistency.} 
Define value match indicator $\delta_k = \mathbb{1}\{v^{\text{gt}}_k = v^{\text{pred}}_k\}$ for key $k$. Compute:
\begin{align}
\text{ValueMatch} &= \frac{1}{|\mathcal{K}^c|} \sum_{k \in \mathcal{K}^c} \delta_k \\
\text{NumericMAE} &= \frac{1}{|\mathcal{N}|} \sum_{k \in \mathcal{N}} |v^{\text{gt}}_k - v^{\text{pred}}_k|
\end{align}
where $\mathcal{N} \subseteq \mathcal{K}^c$ contains numeric keys.

\noindent\textit{Overall score.} 
Combine topology and content with 7:3 weighting:
\begin{equation}
\text{JsonStructSim} = 0.7 \cdot \text{Key-}F_1 + 0.3 \cdot \text{ValueMatch}
\end{equation}

\noindent\textbf{Semantic-Level Metrics.}
We evaluate alignment between generated content and input text prompts using CLIP~\cite{radford2021learning}. For static image generation tasks, we measure CLIP text-image similarity between the input prompt and rendered image. For animation generation tasks, we render the animation into video frames and compute the average CLIP score across all frames.

\begin{figure*}[t]
    \centering
    \includegraphics[width=\linewidth]{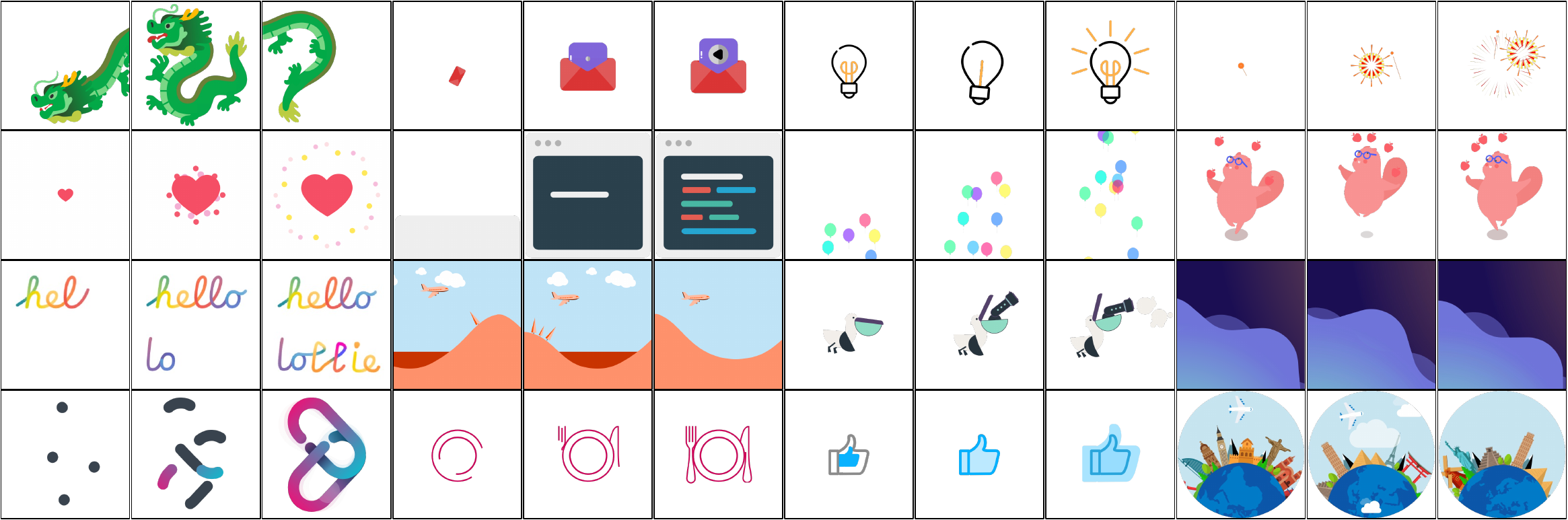}
    \caption{
    Lottie animations generated by LottieGPT.
    }
    \vspace{-3mm}
    \label{fig:comp_animation_text_3x3}
    \vspace{-3mm}
\end{figure*}

\section{Results}

\subsection{Baselines}

For the vector graphics generation task, we compare LottieGPT with state-of-the-art SVG generation methods OmniSVG~\cite{yang2025omnisvg} and StarVector~\cite{rodriguez2025starvector}.
For the vector animation generation task, we compare LottieGPT with state-of-the-art video generation models including Sora 2~\cite{openai2024sora2}, Kling~\cite{kling}, and Veo 3.1~\cite{google2024veo31}.
We also evaluate against state-of-the-art commercial models, including GPT-5~\cite{openai2024gpt5}, Claude Sonnet 4.5~\cite{anthropic2024claude}, Gemini 2.5 Pro~\cite{google2024gemini25pro}, Qwen3-Max~\cite{qwen3technicalreport}, and DeepSeek-V3.1~\cite{deepseekai2024deepseekv3technicalreport}.
Other vector animation generation methods, including LINR-bridge~\cite{gao2025linr} and AniClipart~\cite{wu2025aniclipart}, are based on LiveSketch~\cite{LIVESKETCH}. However, LiveSketch does not support SVG inputs with \textit{Groups} and \textit{Transforms}, both of which are essential components for vector animations. As a result, these methods cannot handle the complex structure of Lottie animations.

\subsection{Training Details}
We follow Qwen2.5-VL's hyperparameters: frozen vision encoder, trainable MLP adapter and LLM, learning rate $1\times10^{-6}$, max sequence length 40K tokens. Each stage trains for 20 epochs on 8×H20 GPUs (140GB each) for ~1 week using bfloat16 and DeepSpeed ZeRO-3.
We expand the vocabulary with 441 Lottie tokens and 35 padding tokens (152,064→152,128, padded to multiples of 64).
Due to GPU constraints, we use 750K MMSVG-icon samples for Stage 1 and 60K Lottie animations for Stage 2.

\subsection{Evaluations and Comparisons}

\noindent\textbf{Static Vector Graphics Task.}
We evaluate static vector graphics generation under two settings: text-only input and text-with-image input. Tab.~\ref{tab:animation_comp} presents comprehensive results on LottieBench. For fair comparison, all baselines including LottieGPT-Stage1 are trained on MMSVG-Icon, the dataset used by OmniSVG.
Experimental results show that LottieGPT achieves superior performance on both text-to-vector and image-to-vector generation tasks, outperforming the current state-of-the-art method OmniSVG across all metrics. This validates the effectiveness of Lottie JSON as a vector graphics representation format. 
We refer readers to the supplementary material for qualitative comparisons.
Following OmniSVG's experimental protocol, we train and evaluate on identical datasets. 
It is important to note that LottieGPT-7B-Stage1 in Tab.~\ref{tab:animation_comp} is trained on only 750K samples from MMSVG-Icon, whereas OmniSVG leverages 2M samples from both MMSVG-Icon and MMSVG-Illustrations datasets.

\noindent\textbf{Vector Animation Task.}
For the vector video generation task, we similarly evaluated using text-only prompts as input and using text prompts with image input. We present the results of all vector animation methods on LottieBench in Tab.~\ref{tab:animation_comp}. Existing SVG-based vector animation methods~\cite{LIVESKETCH,gao2025linr,wu2025aniclipart} are unable to perform inference on the SVGs converted from our LottieAnimation dataset.
Since baseline methods can hardly generate meaningful animations, please refer to the supplementary materials for qualitative comparison results. Fig.~\ref{fig:comp_animation_text_3x3} shows some generation examples from LottieGPT. Experimental results demonstrate that LottieGPT significantly outperforms state-of-the-art video generation methods and few-shot VLM results.

\subsection{Ablation Study}
Tab.~\ref{tab:animation_comp} includes a variant finetuned directly with Lottie JSON without our tokenizer (Finetuned w. Lottie JSON). Due to the frequent occurrence of sequences exceeding 40K tokens, we train only on samples within the 40K token limit, with pretraining on MMSVG-Icon converted to Lottie JSON.
We also evaluate LottieGPT-7B-Stage1 (trained only on MMSVG-Icon) on animation tasks. Despite achieving state-of-the-art performance on static graphics, it performs poorly on animation generation. This validates the effectiveness of our Lottie Tokenizer and Static-to-Dynamic training strategy.

\section{Conclusion}

We present LottieGPT, the first framework for autoregressively generating editable vector animations from text, images, or keyframes. By designing a Lottie Tokenizer for hierarchical primitives and keyframes, constructing a 660K animation dataset, and fine-tuning Qwen2.5-VL via "static-to-dynamic" training, LottieGPT produces resolution-independent, compact, and editable outputs, achieving state-of-the-art performance in both vector animation and SVG generation.

\section*{Acknowledgments}
This work is supported in part by the National Natural Science Foundation of China under contract No. 62171256, in part by the the Guangdong Natural Science Foundation (2026A1515010184).

{
    \small
    \bibliographystyle{ieeenat_fullname}
    \bibliography{main}
}

\clearpage
\setcounter{page}{1}
\maketitlesupplementary

\tableofcontents
\clearpage

\begin{figure*}
    \centering
    \includegraphics[width=1\linewidth]{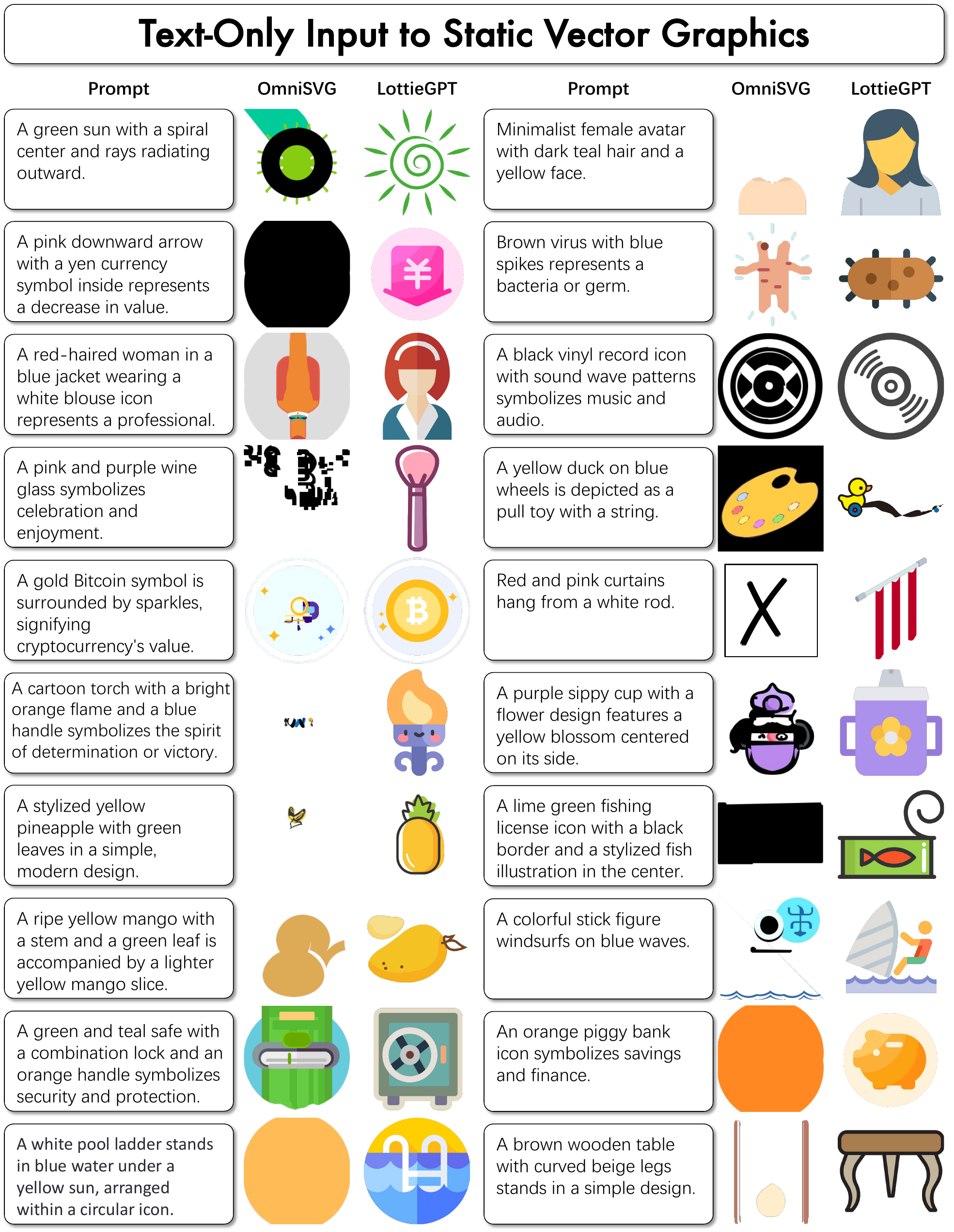}
    \caption{Performance of OmniSVG and LottieGPT on the text-to-vector graphics generation task.}
    \label{fig:comp_text_to_svg}
\end{figure*}

\begin{figure*}
    \centering
    \includegraphics[width=1\linewidth]{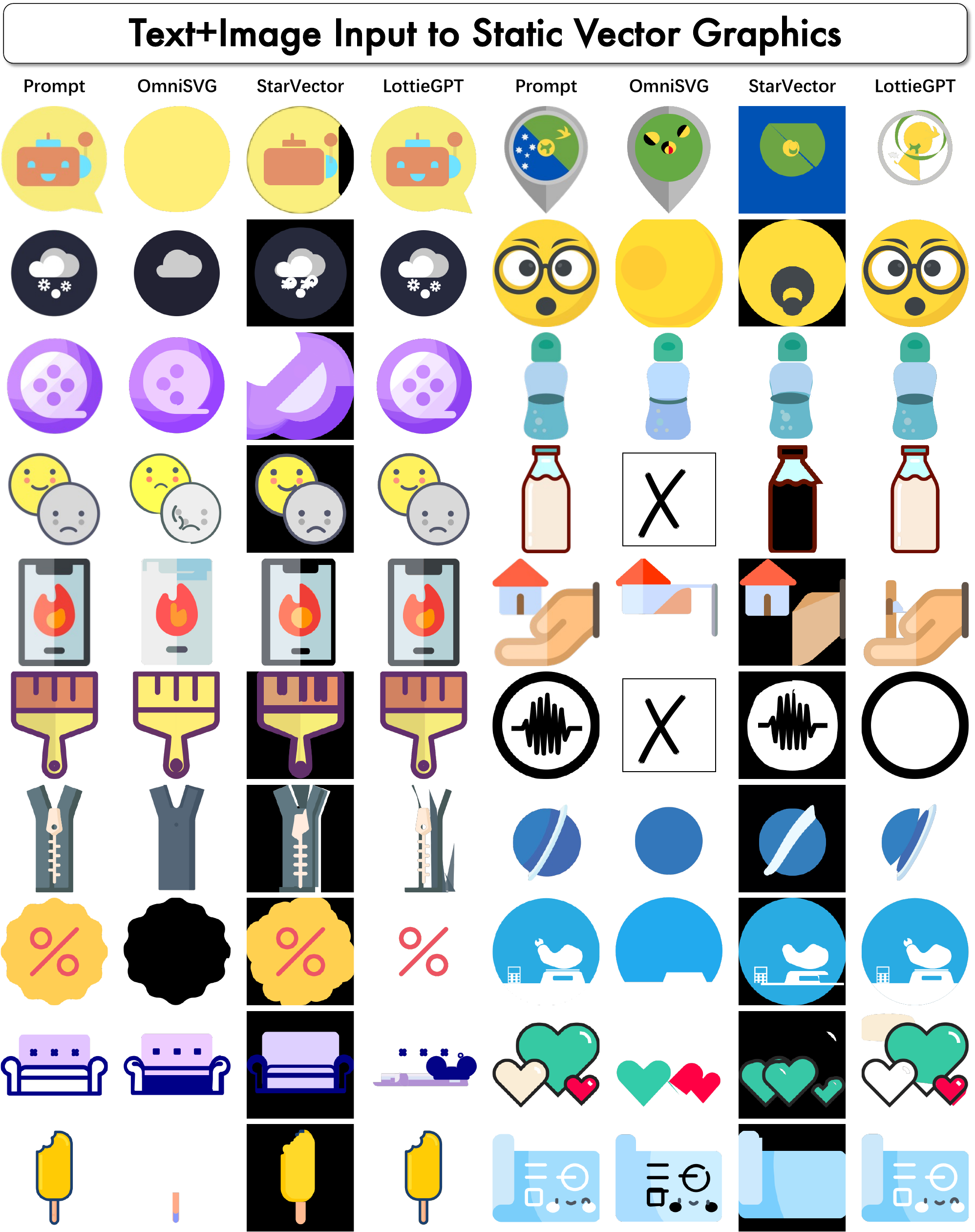}
    \caption{Performance of OmniSVG, StarVector, and LottieGPT on the image-to-vector graphics generation task.}
    \label{fig:comp_text_image_to_svg}
\end{figure*}

\begin{figure*}
    \centering
    \includegraphics[width=1\linewidth]{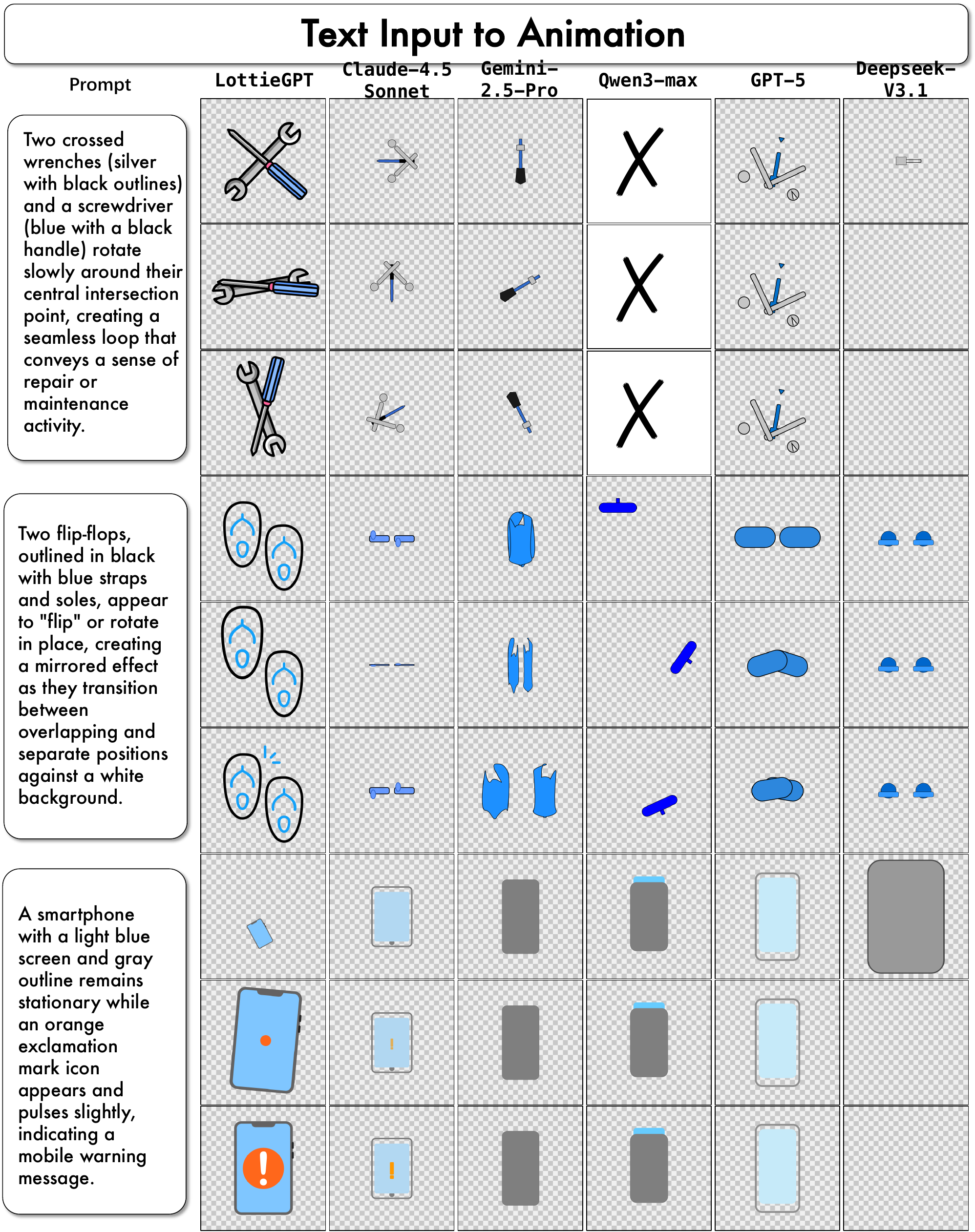}
    \caption{For the Text-to-Animation task, all LLM baselines were provided with identical 3-shot examples. 
\textit{\xmark} indicates that no renderable Lottie JSON was obtained even after the fifth attempt.
\textbf{More results on vector animation can be found in the supplementary video and on the project website.}
}
    \label{fig:comp_text_to_animation}
\end{figure*}

\begin{figure*}
    \centering
    \vspace{-3mm}
    \includegraphics[width=1\linewidth]{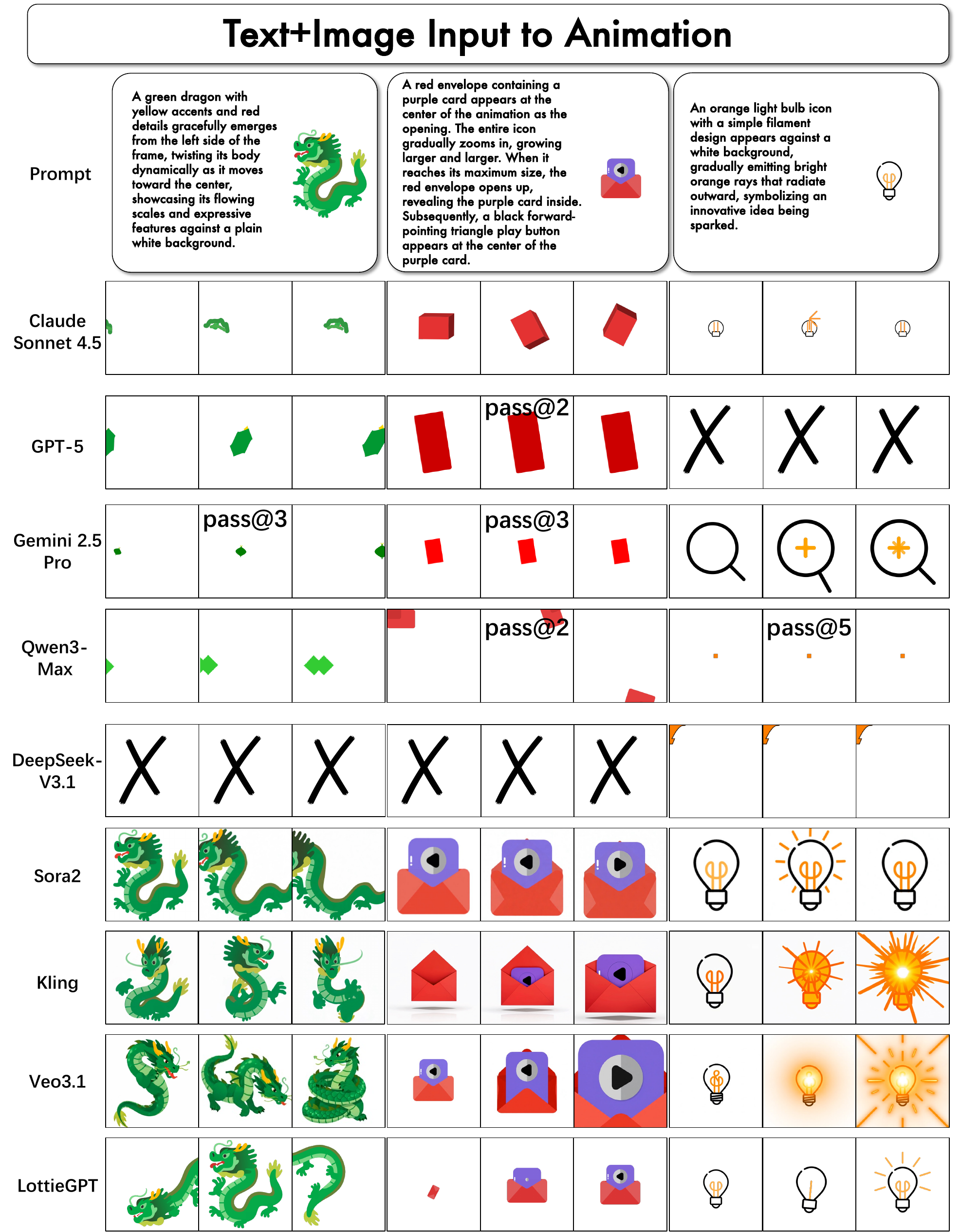}
    \caption{Using Text+Image as input to generate animations. Few-shot refers to providing three description-Lottie JSON data pairs. Except for Deepseek which does not support image input, all other methods use a single image and text description as input. pass@x indicates that x attempts were required to generate a renderable Lottie JSON. \textit{\xmark} indicates that no renderable Lottie JSON was obtained even after the fifth attempt.}
    \label{fig:comp_animation_text_3x3_full}
\end{figure*}

\begin{figure*}
    \centering
    \vspace{-3mm}
    \includegraphics[width=1\linewidth]{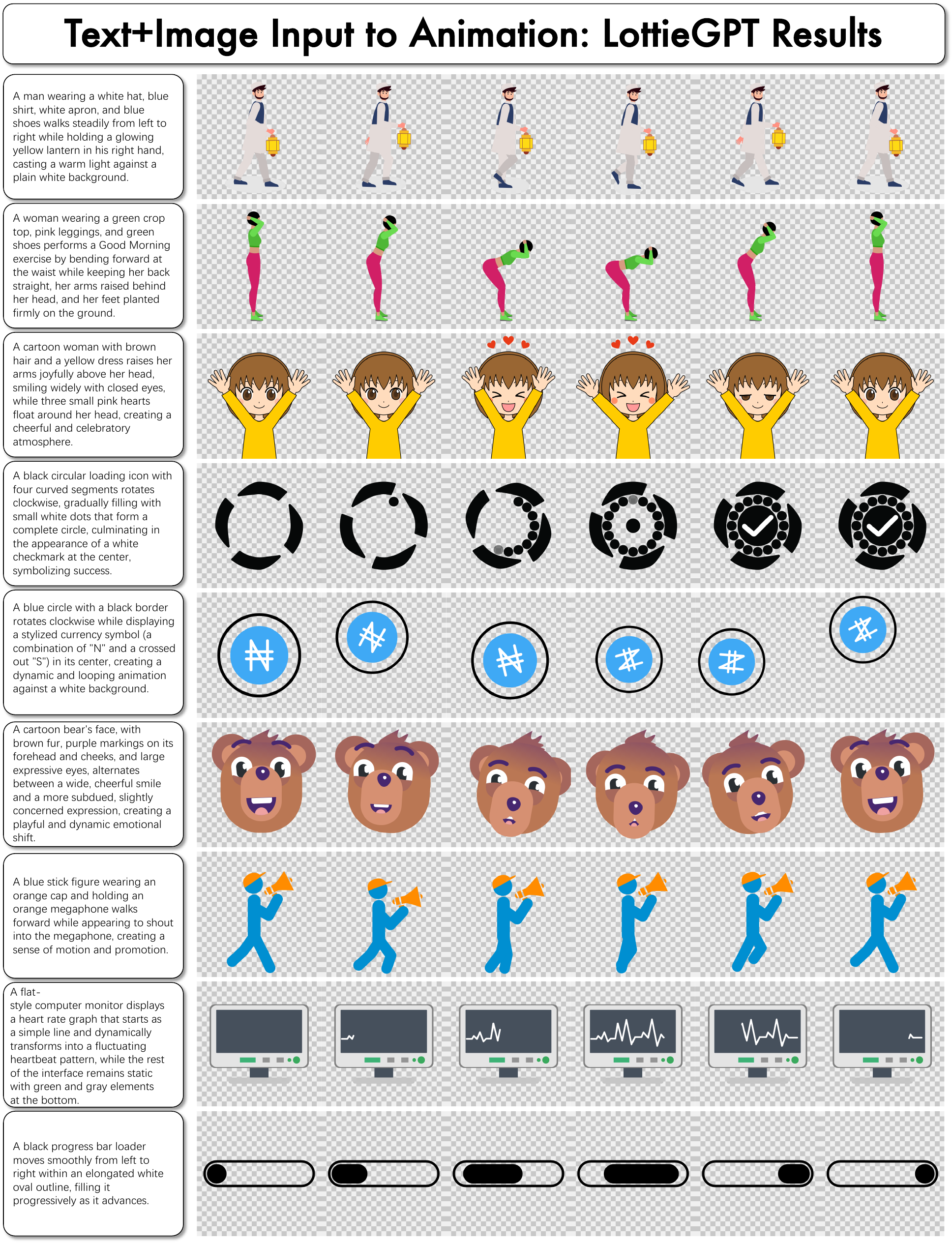}
    \caption{LottieGPT results on in-the-wild text-only inputs.}
    \label{fig:text_to_animation_lottiegpt_demos}
\end{figure*}

\begin{figure*}
    \centering
    \includegraphics[width=1\linewidth]{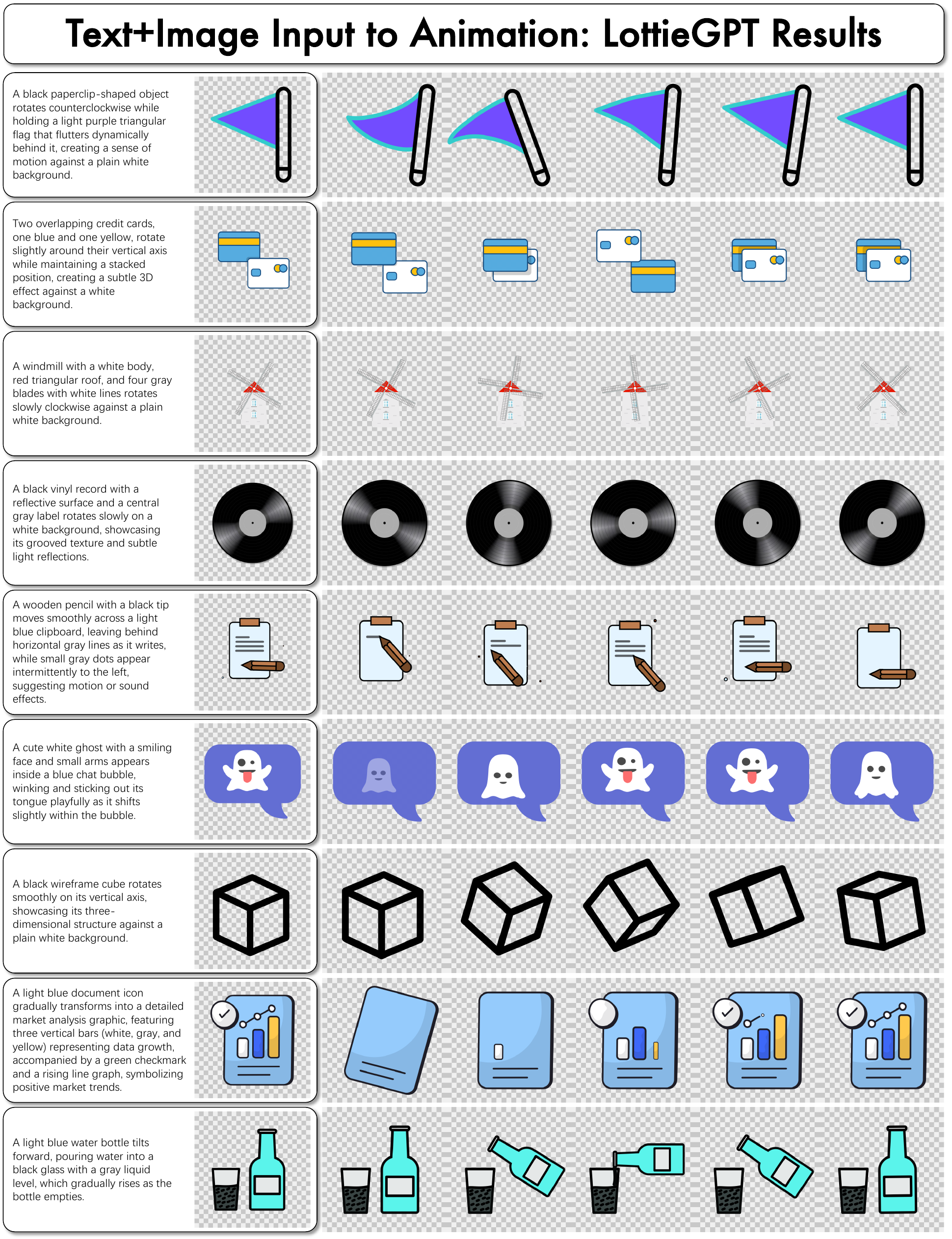}
    \caption{LottieGPT results on in-the-wild text-image inputs.}
    \label{fig:textimage_to_animation_lottiegpt_demos}
\end{figure*}

\section{Comparison of the Tokenizer}

To demonstrate the efficiency of our proposed Lottie tokenizer, we compare it with existing tokenization approaches across different datasets. We evaluate the compression performance on both MMSVG-icon and LottieAnimation datasets using four tokenization methods: QwenVL tokenizer, OmniSVG tokenizer, our Lottie tokenizer, and Lottie tokenizer with numeric quantization.

\subsection{Token Count Comparison}
As shown in Tab.~\ref{tab:tokenizer_comparison}, our Lottie tokenizer achieves superior compression ratios compared to QwenVL and OmniSVG tokenizers.
On the MMSVG-icon dataset, our tokenizer reduces the average token count from 2.6k (QwenVL w. SVG) to 1.3k, achieving a 50\% compression ratio.
Through numerical quantization, this ratio is further improved to 15.4\%, with the average token count reduced to 0.4k. (Note that the OmniSVG tokenizer already incorporates numerical quantization.)
For the more complex LottieAnimation dataset, this advantage becomes even more pronounced. Our tokenizer achieves a 63.3\% compression ratio (17.4k tokens vs. QwenVL's 27.5k tokens), and with quantization, the compression ratio reaches 24\% (6.6k tokens). The significant reduction in token count not only accelerates training and inference but also enables the model to handle longer and more complex animations within the same context window. Unlike the difficulty-based Lottie data segmentation by token count mentioned in the main text, the average token count calculation here includes numerical values to provide a fairer comparison with QwenVL and OmniSVG.

\begin{table}[htbp]
\centering
\caption{Tokenizer comparison on MMSVG-icon and LottieAnimation datasets. Our Lottie tokenizer achieves significantly better compression ratios while maintaining generation quality.}
\label{tab:tokenizer_comparison}
\resizebox{\columnwidth}{!}{
\begin{tabular}{l|ccc|cc}
\toprule
\multirow{2}{*}{\textbf{Tokenizer}} & \multicolumn{3}{c|}{\textbf{Avg. File Size}} & \multicolumn{2}{c}{\textbf{Tokens}} \\
\cmidrule(lr){2-4} \cmidrule(lr){5-6}
& \shortstack{\textbf{PNG/}\\\textbf{MP4}} & \shortstack{\textbf{SVG/}\\\textbf{JSON}} & \shortstack{\textbf{Comp.}\\\textbf{Ratio}} & \textbf{Avg.} & \shortstack{\textbf{Comp.}\\\textbf{Ratio}} \\
\midrule
\rowcolor{orange!20}
\multicolumn{6}{c}{\textit{\textbf{MMSVG-icon Dataset}}} \\
\midrule
SVG Code & 5.46 KB & 2.74 KB & 50.2\% & N/A & N/A \\
Lottie Json & 5.46 KB & 2.16 KB & 39.6\% & N/A & N/A \\
\midrule
QwenVL w. SVG & N/A & N/A & N/A & 2.6 k & 100\% \\
OmniSVG & N/A & N/A & N/A & 2.2 k & 84.6\% \\
\hline
QwenVL w. JSON & N/A & N/A & N/A & 1.6 k & 61.5\% \\
Lottie & N/A & N/A & N/A & 1.3 k & 50.0\% \\
\rowcolor{gray!20}
\textbf{Lottie w. Quant.} & N/A & N/A & N/A & \textbf{0.4 k} & \textbf{15.4\%} \\
\midrule
\rowcolor{orange!20}
\multicolumn{6}{c}{\textit{\textbf{LottieAnimation Dataset}}} \\
\midrule
Original & 194.11 KB & 60.72 KB & 31.3\% & N/A & N/A \\
Optimized & 194.11 KB & 39.87 KB & 20.5\% & N/A & N/A \\
\hline
QwenVL & N/A & N/A & N/A & 27.5 k & 100\% \\
Lottie & N/A & N/A & N/A & 17.4 k & 63.3\% \\
\rowcolor{gray!20}
\textbf{Lottie w. Quant.} & N/A & N/A & N/A & \textbf{6.6 k} & \textbf{24.0\%} \\
\bottomrule
\end{tabular}
}
\end{table}

\subsection{Animation File Size Comparison}
We find that using Lottie JSON as the representation for vector graphics and vector animations yields higher compression ratios. For example, as shown in Tab.~\ref{tab:tokenizer_comparison}, on the MMSVG-icon dataset, the average size of original SVG files is 2.74KB, which is 50.2\% of the average PNG image size of 5.46KB. However, vector graphics stored in Lottie JSON format have an average size of only 2.16KB, merely 39.6\% of the original PNG image size.
This can also be reflected in token counts. For example, using the QwenVL default tokenizer on the MMSVG-icon dataset with SVG files as training instructions, the average token length is 2.6k (QwenVL w. SVG), while using Lottie JSON files as training instructions, the average token length is only 1.6k, which is 61.5\% of the former, with no difference in rendering results between the two.
On the LottieAnimation dataset, the average size of MP4 files rendered from original Lottie animations is 194.11KB, while animations saved in Lottie JSON format have an average size of 60.72KB, achieving a compression ratio of 31.3\%. After our simplification process, the average Lottie JSON size is further reduced to 39.87KB, improving the compression ratio to 20.5\%.
This demonstrates that using Lottie JSON for representing vector graphics and vector animations achieves higher compression ratios than SVG without sacrificing rendering quality.

\subsection{Support of Numeric Quantization}
In the Lottie JSON simplification process, we simplify the numerical components by compressing floating-point values to four significant digits. This approach significantly reduces the size of Lottie JSON without sacrificing rendering quality.
It is important to note that the current version of LottieGPT presented in the main text was trained \textit{without} numerical quantization to ensure maximum fidelity.
However, our experiments show that models trained with quantized tokens achieve comparable performance while being more efficient.
Inference is performed on a single NVIDIA H20 GPU (140GB), 
with LottieGPT-7B achieving a generation speed of approximately 60 tokens per second.

\section{Static Vector Graphics Task Results}
We compared our method with state-of-the-art SVG generation models, including StarVector~\cite{rodriguez2025starvector} and OmniSVG~\cite{yang2025omnisvg}, conducting experiments with both text input and image input. Qualitative results are shown in Fig.\ref{fig:comp_text_to_svg} and Fig.\ref{fig:comp_text_image_to_svg}.
We randomly sampled examples from LottieBench. As illustrated in the figures, although LottieGPT may still fail on some complex cases, it is capable of generating static vector graphics for the vast majority of scenarios. Notably, our current version was trained on only 1/3 of the training data used by OmniSVG.

\begin{figure*}[t]
    \centering
    \includegraphics[width=\linewidth]{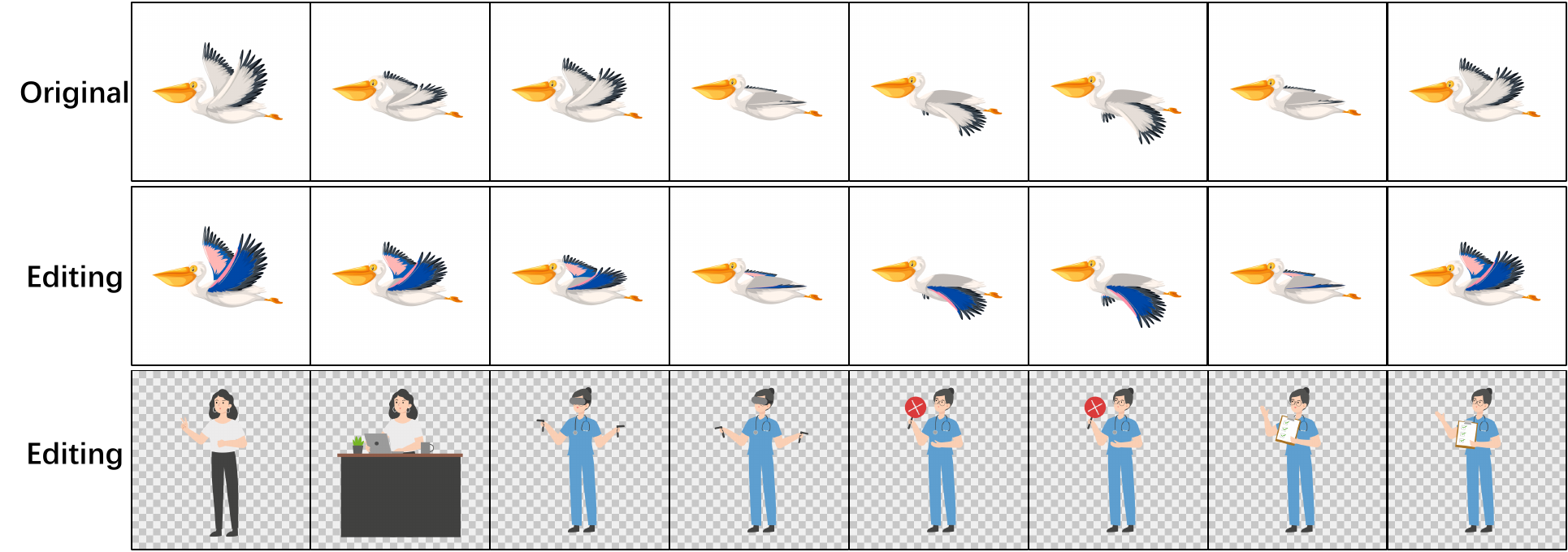}
    \caption{A manually edited Lottie animation where we modified the wing color using LottieLab.}
    \label{fig:editing}
\end{figure*}

\begin{figure}[t]
    \centering
    \includegraphics[width=1\linewidth]{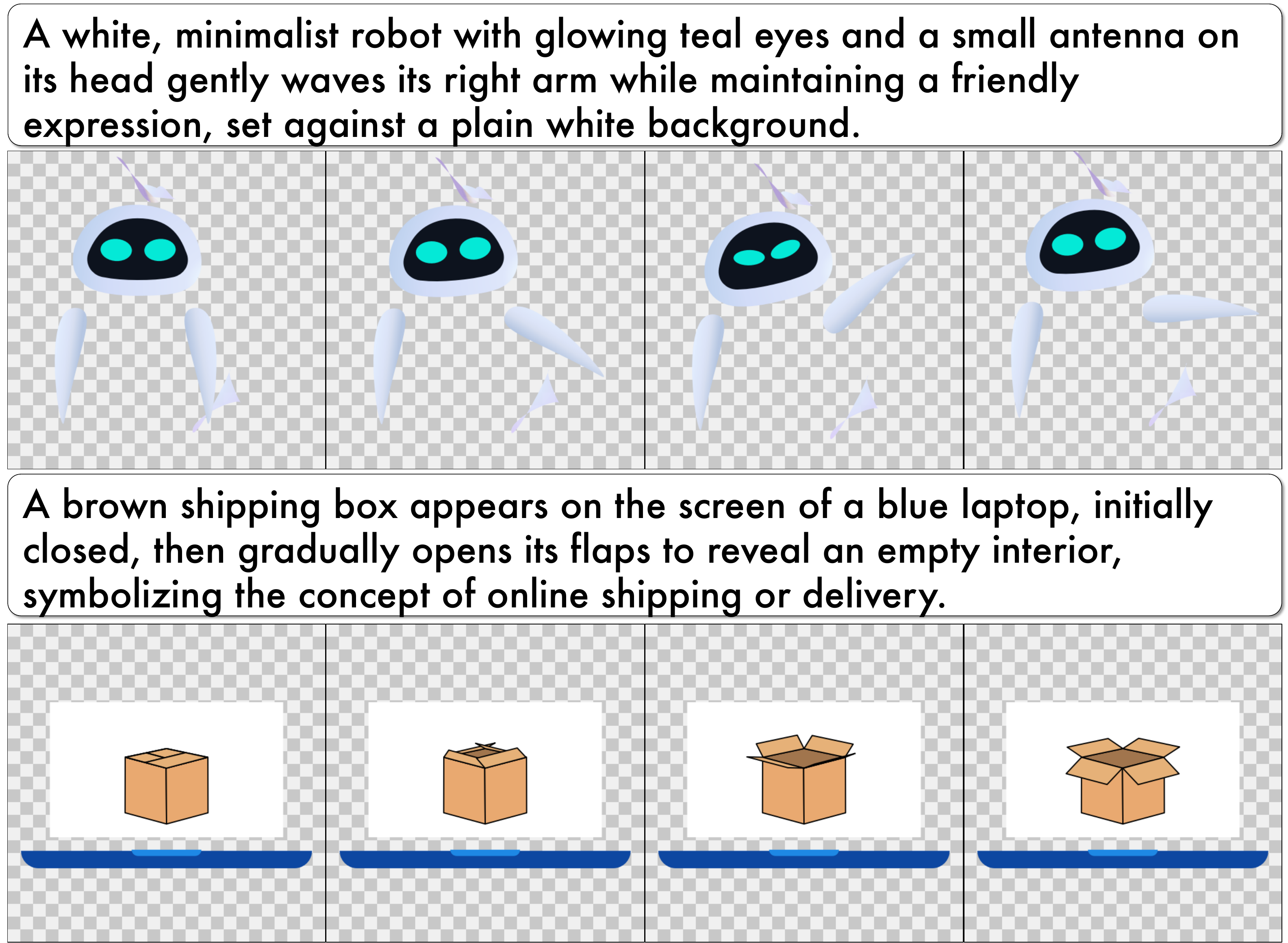}
    \caption{LottieGPT may still generate cases that are renderable but visually inconsistent with expectations, typically manifesting as extraneous or missing shapes relative to the intended design.}
    \label{fig:failure_case}
    \vspace{-2mm}
\end{figure}

\section{Vector Animation Task Results}

All baselines were invoked through their official APIs, and we selected cases where the majority of baselines could successfully render. 
Comparison results are presented in Fig.~\ref{fig:comp_text_to_animation} and Fig.~\ref{fig:comp_animation_text_3x3_full}.
Since zero-shot LLM / VLM cannot generate any valid renderable Lottie JSON files, we do not present these results in our quantitative and qualitative evaluations, and only show the few-shot LLM / VLM results. 

In our experimental results, most LLM-generated outputs failed to render properly.
We present more LottieGPT results on in-the-wild inputs in Fig.~\ref{fig:text_to_animation_lottiegpt_demos} and Fig.~\ref{fig:textimage_to_animation_lottiegpt_demos}.

\textbf{Please refer to the supplementary materials including the project website and demo video for more animation generation comparison results.}

\section{Examples of Editing}
Lottie animations can be created and edited using various professional tools. The most common workflow involves using Adobe After Effects~\footnote{~\url{https://adobe.com/products/aftereffects.html}} with the Bodymovin plugin\footnote{\url{https://aescripts.com/bodymovin/}} to export animations as Lottie JSON format. For online editing, LottieLab~\footnote{\url{https://www.lottielab.com/?home}} and LottieFiles\footnote{\url{https://lottiefiles.com/lottie-editor}} provides a comprehensive ecosystem including animation editors, preview tools, and asset libraries. Additionally, tools such as Haiku Animator\footnote{\url{https://www.haikuanimator.com/}} and Cavalry\footnote{\url{https://cavalry.scenegroup.co/}} also support direct creation and export of Lottie animations. 
To provide a more intuitive illustration of the editability and flexibility advantages of Lottie Animation over traditional raster videos, we present an example from the dataset in Fig.~\ref{fig:editing} with various editing results.

\section{Failure Cases}
Some failure cases of LottieGPT are shown in Fig.~\ref{fig:comp_text_image_to_svg} and Fig.~\ref{fig:failure_case}.
The Valid Rate in Tab. 2 of the main text refers to the rendering success rate rather than the exact match rate. Most Lottie JSON files generated by VLMs cannot be rendered properly (they do not conform to the standard Lottie JSON format).
Rendering failures in LottieGPT typically arise from two scenarios.
First, the presence of non-numeric content, such as unexpected characters, in numerical values during Lottie token generation causes detokenization to fail. 
Second, when the generated Lottie token sequence exceeds the maximum context length, it becomes truncated, preventing successful detokenization and valid Lottie JSON generation.
Similarly, the rendering failures of OmniSVG shown in Fig.~\ref{fig:comp_text_image_to_svg} are attributed to the generation of unexpected characters that cause detokenization failures.

\begin{figure*}[t]
    \centering
    \includegraphics[width=0.76\linewidth]{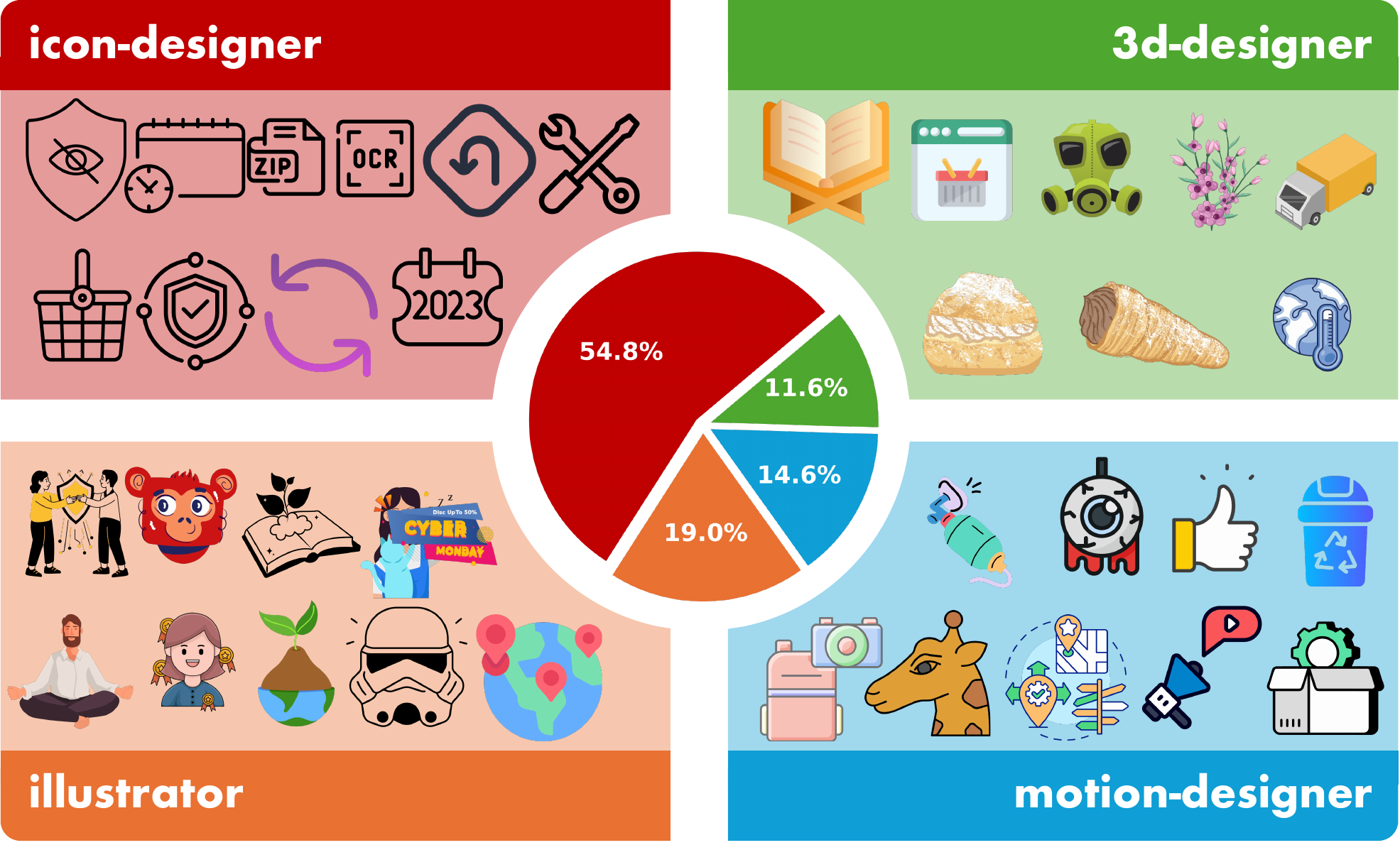}
    \caption{Representative examples from the four categories in LottieSVG-10M dataset. From top to bottom: icon-designer, illustrator, motion-designer, and 3d-designer. Each category exhibits distinct visual characteristics and complexity levels.}
    \label{fig:svg_data_examples}
    \vspace{-2mm}
\end{figure*}

\section{Lottie Dataset}\label{supp_lottiedataset}

We introduce three comprehensive datasets for vector graphics and animation generation: LottieSVG-10M, LottieImage-15M, and LottieAnimation-660K. These datasets provide paired data of vector code, textual descriptions, and rendered images/videos, enabling multimodal learning for vector graphics generation.
\textbf{Please refer to the supplementary materials for dataset examples.}

\subsection{Details of Data Curation Pipeline}

While pixel-based visual data has become abundant with the rapid development of generative models for images and videos, and several large-scale static vector graphics datasets (e.g., SVG) are widely used, existing resources focus almost exclusively on static content, lacking systematically curated large-scale vector animation datasets.

Due to SVG's limited support for complex timeline-based animations, obtaining high-quality vector animations directly is challenging. We therefore turn to the After Effects (AE) ecosystem, leveraging the Bodymovin plugin\footnote{\url{https://aescripts.com/bodymovin/}} to export AE animations into Lottie JSON format. As a lightweight, cross-platform vector animation representation, Lottie supports dynamic properties such as keyframes, path morphing, and color gradients in JSON format, making it well-suited for autoregressive model training.

We collect a large number of AE source files from public resources and convert them to Lottie format via an automated pipeline. Additionally, we integrate over 15 million static vector graphics, uniformly converted to Lottie representation, to support a progressive ``static-first, then dynamic'' training strategy. To enable multimodal generation, we use BLIP-2~\cite{li2023blip} to generate text descriptions for static vector graphics, and employ Qwen2.5-VL 32B~\cite{bai2025qwen2} to produce detailed, temporally aligned descriptions after rendering vector animations into videos. Fig.~\ref{fig:datacurationpipe} illustrates our dataset construction pipeline and sample examples.

During dataset construction, we apply rigorous filtering and standardization. First, we remove all animations with rendering failures or visual anomalies (e.g., blank frames, misalignment, flickering). Second, we simplify Lottie JSON structures by removing redundant fields that do not affect rendering (e.g., comments, unused layer properties, debug metadata), and unify version formats and key path representations to ensure data consistency and training stability. The right side of Fig.~\ref{fig:pipeline} shows the word cloud of text descriptions and file size distribution for the LottieAnimation dataset. Our simplification reduces sequence length by approximately 34\% without affecting animation rendering quality.

LottieImage and LottieAnimation cover diverse graphic design assets including vector icons, infographics, animated illustrations, cartoon character animations, and UI motion effects, exhibiting rich semantic and stylistic variation. We present the first systematically curated, large-scale dataset of paired vector graphics and animations in \textbf{Lottie format}.
Tab.~\ref{tab:dataset_comparison} compares our datasets with existing vector graphics datasets.
This dataset provides a novel structured representation foundation for vector generation, multimodal understanding, and text-to-animation synthesis, advancing generative models toward scalable, editable, and lightweight dynamic content.

\subsection{LottieSVG-10M Dataset}

We collect and filter 10 million unique SVG images from the internet, ensuring no overlap with existing SVG datasets. Following the annotation methodology of OmniSVG, we employ BLIP-2 with the instruction template shown in Fig.~\ref{fig:instruction_templates_svg_cap} to generate textual descriptions for each SVG image. The LottieSVG-10M dataset provides triplets of \textit{(SVG code, Lottie Json, text description, rendered PNG image)}.

\textbf{Category Distribution.} The dataset comprises four main categories based on creator designations, as shown in Fig.~\ref{fig:svg_data_examples}. Icon-designer content dominates with 54.82\%, followed by illustrator (18.96\%), motion-designer (14.64\%), and 3d-designer (11.58\%). Representative examples from each category are visualized in Fig.~\ref{fig:svg_data_examples}.

\begin{figure}[h]
    \centering
    \includegraphics[width=0.95\columnwidth]{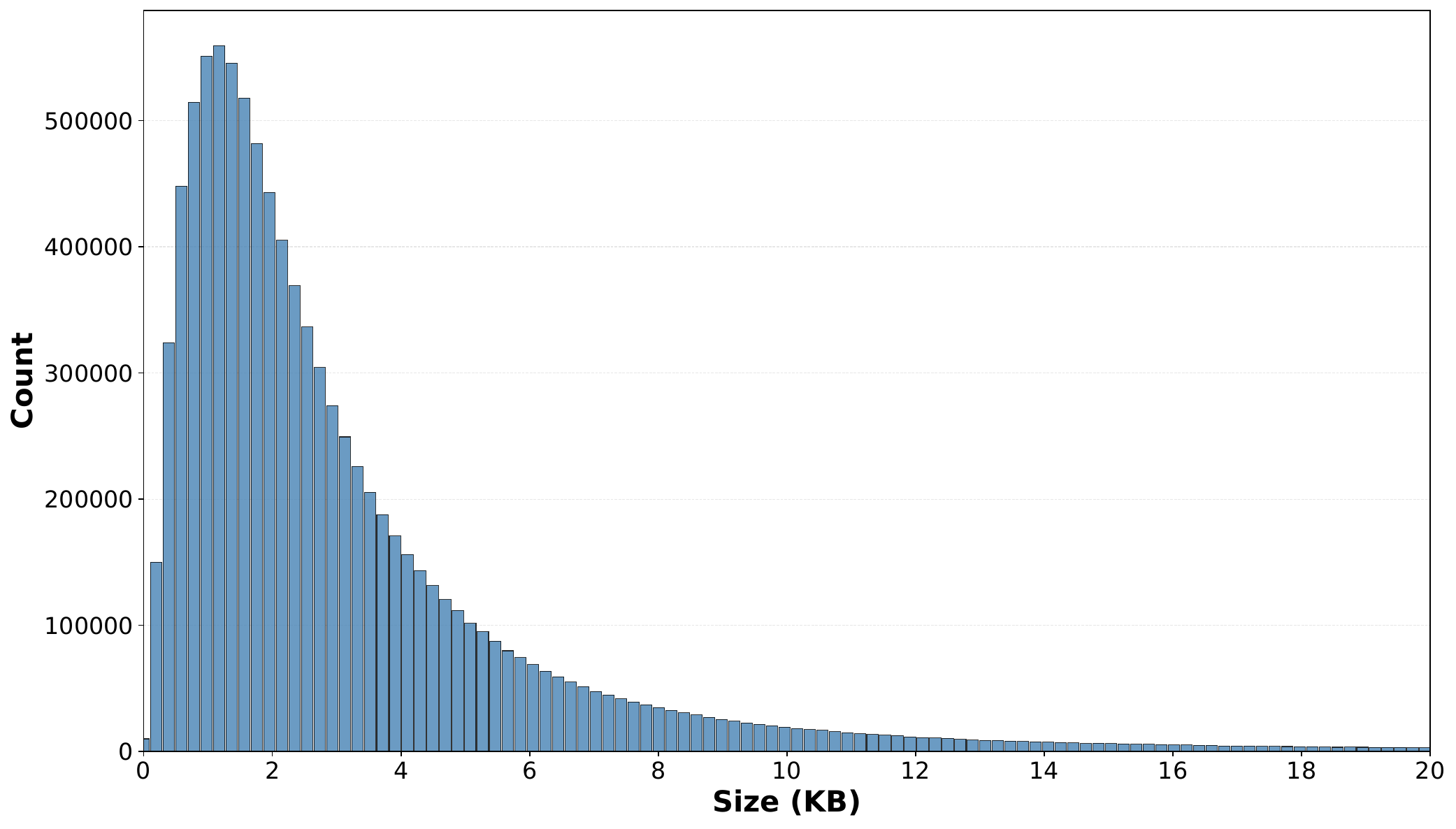}
    \caption{File size distribution of LottieSVG-10M dataset in the 0-20KB range. The distribution shows a mean of 4.65KB and median of 2.25KB, demonstrating the compact nature of SVG representations.}
    \label{fig:svg_size_distribution}
    \vspace{-4mm}
\end{figure}

\textbf{File Size Distribution.} Fig.~\ref{fig:svg_size_distribution} shows the file size distribution of SVG images in the 0-20KB range, which covers the majority of the dataset. The distribution exhibits a right-skewed pattern with a mean size of 4.65KB and median of 2.25KB, indicating that most SVG files are compact and efficient for storage and transmission.

\begin{figure}[h]
    \centering
    \includegraphics[width=0.95\columnwidth]{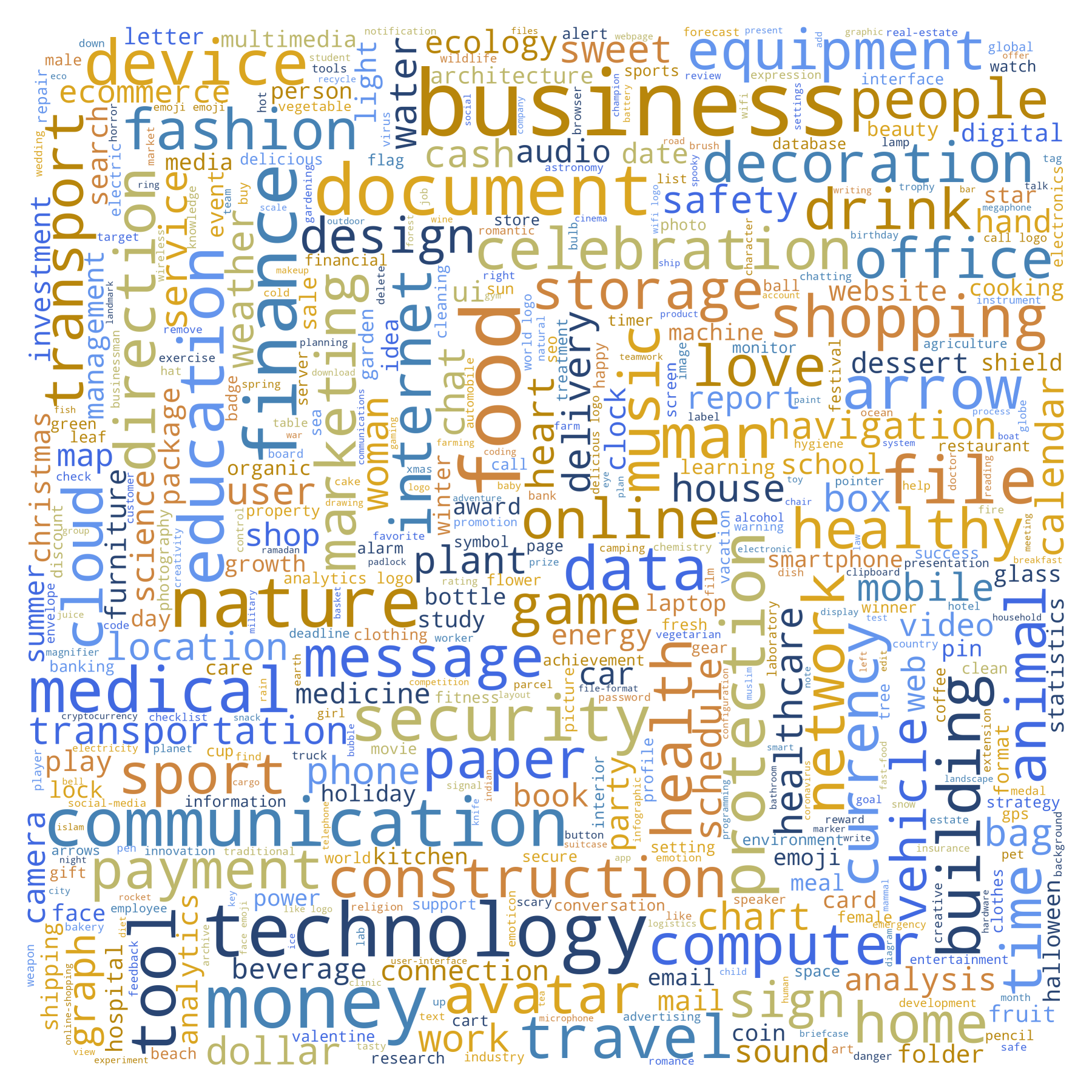}
    \caption{Word cloud visualization of tag distribution in LottieSVG-10M dataset. The size of each word corresponds to its frequency, with top tags including business, food, money, technology, and finance.}
    \label{fig:svg_tag_wordcloud}
    \vspace{-4mm}
\end{figure}

\textbf{Tag Distribution.} To analyze the semantic content of our dataset, we visualize the tag distribution as a word cloud in Fig.~\ref{fig:svg_tag_wordcloud}. The most frequent tags include ``business'' (535,098), ``food'' (407,321), ``money'' (316,625), ``technology'' (314,205), and ``finance'' (288,782), reflecting the diverse application domains covered by our dataset.

\begin{figure}[h!]
\centering
\begin{minipage}{0.99\columnwidth}
\vspace{0mm}    
\centering
\begin{tcolorbox} 
    \centering
    \small
    \hspace{-6mm}
\begin{itemize}[leftmargin=0mm]
\setlength{\itemsep}{2pt}
    \item \textbf{SVG data Captioning:} You are a helpful assistant. Your task is to describe this image in a single sentence, including the object, its color, and its overall arrangement. For example: “Yellow cheers with glasses of alcohol drinks.” / “Heart emojis represent love on Valentine’s Day.”
\end{itemize}
\end{tcolorbox}
\vspace{-4mm}
\caption{The instruction for SVG data captioning.}
\label{fig:instruction_templates_svg_cap}
\end{minipage}
\vspace{-4mm}
\end{figure}

\subsection{LottieImage-15M Dataset}

We construct the largest vector graphics dataset to date by combining LottieSVG-10M, SVG-Stack-2M, OmniSVG-2M, and SVGX-1M, totaling 15 million SVG images with detailed textual annotations. 

To enable Lottie-based generation, we develop a conversion pipeline that transforms SVG static images into Lottie static images. The conversion process includes:
\begin{itemize}[leftmargin=*]
    \item Parsing SVG elements and attributes
    \item Mapping SVG primitives to Lottie shape layers
    \item Converting coordinate systems and transformations
    \item Validating rendering consistency
\end{itemize}

We filter out samples where the rendered output differs between SVG and Lottie formats, ensuring high-quality paired data. The LottieImage-15M dataset provides quadruplets of \textit{(SVG code, Lottie JSON code, text description, rendered PNG image)}.

\subsection{LottieAnimation-660K Dataset}

The LottieAnimation-660K dataset contains 671,121 Lottie animations with comprehensive temporal annotations. The dataset comprises a total of 70,261,325 frames, spanning 608 hours of animation content. On average, each animation contains 104.7 frames with a median of 90 frames, ranging from 1 to 3,600 frames. In terms of duration, the animations average 3.26 seconds with a median of 3.00 seconds, ranging from 0.02 to 91.40 seconds.

\textbf{Duration Distribution.} As shown in Tab.~\ref{tab:duration_distribution}, the animation durations exhibit a concentrated distribution pattern. The majority of animations (70.55\%) fall within the 2-5 second range, which is typical for UI animations and micro-interactions. Short animations (1-2s) account for 10.29\%, while very short animations (0-1s) represent only 1.51\%. Longer animations (5-10s) comprise 16.19\%, and ultra-long animations exceeding 10 seconds constitute merely 1.46\% of the dataset, indicating a focus on concise, purposeful animations rather than extended narrative sequences.

\begin{table}[h]
\centering
\caption{Duration distribution of LottieAnimation-660K dataset.}
\label{tab:duration_distribution}
\resizebox{0.95\columnwidth}{!}{
\begin{tabular}{lccc}
\toprule
\textbf{Duration Range} & \textbf{Count} & \textbf{Percentage} & \textbf{Total Duration} \\
\midrule
0-1s (Very short) & 10,107 & 1.51\% & 1.98 hours \\
1-2s (Short) & 69,075 & 10.29\% & 27.35 hours \\
2-3s (Medium) & 228,808 & 34.09\% & 135.76 hours \\
3-5s (Long) & 244,724 & 36.46\% & 234.14 hours \\
5-10s (Very long) & 108,633 & 16.19\% & 176.93 hours \\
10s+ (Ultra long) & 9,774 & 1.46\% & 31.90 hours \\
\bottomrule
\end{tabular}
}
\end{table}

\textbf{Simplification and Optimization.} To reduce token count while maintaining visual quality, we apply a multi-step optimization process:
\begin{enumerate}[leftmargin=*]
    \item \textbf{Expression removal:} Remove After Effects expressions that are not supported in standard Lottie players
    \item \textbf{Field pruning:} Remove unused fields including \texttt{nm} (name), \texttt{mn} (match name), \texttt{hd} (hidden), \texttt{ix} (index), and \texttt{cix} (undocumented)
    \item \textbf{Precision reduction:} Round numerical values to 4 significant digits (excluding color values)
    \item \textbf{Color compression:} Convert color values to hexadecimal representation
    \item \textbf{Metadata update:} Update Lottie JSON keys for backward compatibility
\end{enumerate}

We utilize the \texttt{lottie-optim}\footnote{\url{https://github.com/levibuzolic/lottie-optim}} tool for automated optimization.
As shown in Tab.~\ref{tab:tokenizer_comparison}, this simplification approach reduces the average Lottie JSON length by approximately 34\% (from 60.72 KB to 39.87 KB) on the LottieAnimation dataset while maintaining rendering quality with no perceptible visual degradation.

\begin{figure}[h!]
\centering
\begin{minipage}{0.99\columnwidth}
\vspace{0mm}    
\centering
\begin{tcolorbox} 
    \centering
    \small
    \hspace{-6mm}
\begin{itemize}[leftmargin=0mm]
\setlength{\itemsep}{2pt}
    \item \textbf{Lottie Animation data Captioning:} You are a helpful assistant. Your task is to describe this animation in a single sentence, including the objects involved, their colors, positions, actions, and overall layout. This description should enable a designer unfamiliar with the animation to understand its content, style, and structure and create a similar one. For example: "A blue arrow moves from left to right with a faint blue flicker effect." "A red character rotates around a circular path, waving a flag in hand." The title of this video is '\{\textit{video\_title}\}', please describe this animation in a single sentence, including the objects involved, their colors, positions, actions, and overall layout. Be sure to include any movements or dynamic changes occurring in the video.
\end{itemize}
\end{tcolorbox}
\vspace{-2mm}
\caption{The instruction for Lottie animation data captioning.}
\label{fig:instruction_templates_animation_cap}
\end{minipage}
\vspace{-4mm}
\end{figure}

\textbf{Annotation.} We employ Qwen2.5-VL to generate textual descriptions for Lottie animations using the instruction template shown in Fig.~\ref{fig:instruction_templates_animation_cap}. The annotation process considers both spatial content and temporal dynamics, producing descriptions that capture motion patterns, timing, and visual effects.

\subsection{Instruction Templates}
We design task-specific instruction templates for different generation scenarios. Figures~\ref{fig:instruction_text_to_svg} through~\ref{fig:instruction_textvideo_to_animation} present the complete set of instruction templates used in our framework:

\begin{itemize}[leftmargin=*]
    \item \textbf{Static image generation:} Text-to-SVG (Fig.~\ref{fig:instruction_text_to_svg}) and text+image-to-SVG (Fig.~\ref{fig:instruction_textimage_to_svg})
    \item \textbf{Animation generation:} Text-to-Lottie (Fig.~\ref{fig:instruction_text_to_animation}), text+image-to-Lottie (Fig.~\ref{fig:instruction_textimage_to_animation}), and text+video-to-Lottie (Fig.~\ref{fig:instruction_textvideo_to_animation})
\end{itemize}
All generation instructions emphasize the use of compressed token format with special tokens, explicitly prohibiting direct JSON output to ensure compatibility with our tokenization scheme.

\begin{figure}[h!]
\centering
\begin{minipage}{0.99\columnwidth}
\vspace{0mm}    
\centering
\begin{tcolorbox} 
    \centering
    \small
    \hspace{-6mm}
\begin{itemize}[leftmargin=0mm]
\setlength{\itemsep}{2pt}
    \item \textbf{Text to Lottie Image Instruction:} You are a helpful Animation Generator. You output animations in compressed token format using special tokens. You never output JSON format. Generate an static animation based on this description: "\{\textit{caption}\}"

\end{itemize}
\end{tcolorbox}
\vspace{-2mm}
\caption{The instruction for text to Lottie image generation.}
\label{fig:instruction_text_to_svg}
\end{minipage}
\vspace{-4mm}
\end{figure}

\begin{figure}[h!]
\centering
\begin{minipage}{0.99\columnwidth}
\vspace{0mm}    
\centering
\begin{tcolorbox} 
    \centering
    \small
    \hspace{-6mm}
\begin{itemize}[leftmargin=0mm]
\setlength{\itemsep}{2pt}
    \item \textbf{Text and Image to Lottie Image Instruction:} You are a helpful Animation Generator. You output animations in compressed token format using special tokens. You never output JSON format. Generate an static animation based on the given image and this description: "\{\textit{caption}\}"

\end{itemize}
\end{tcolorbox}
\vspace{-2mm}
\caption{The instruction for text and image to Lottie image generation.}
\label{fig:instruction_textimage_to_svg}
\end{minipage}
\end{figure}

\begin{figure}[h!]
\centering
\begin{minipage}{0.99\columnwidth}
\vspace{0mm}    
\centering
\begin{tcolorbox} 
    \centering
    \small
    \hspace{-6mm}
\begin{itemize}[leftmargin=0mm]
\setlength{\itemsep}{2pt}
    \item \textbf{Text to Lottie Animation Instruction:} You are a helpful Lottie animation Generator. You output animations in compressed token format using special tokens. You never output JSON format. Generate Lottie animation based on the description: "\{\textit{caption}\}"
\end{itemize}
\end{tcolorbox}
\vspace{-2mm}
\caption{The instruction for text to Lottie animation generation.}
\label{fig:instruction_text_to_animation}
\end{minipage}
\vspace{-2mm}
\end{figure}

\begin{figure}[h!]
\centering
\begin{minipage}{0.99\columnwidth}
\vspace{0mm}    
\centering
\begin{tcolorbox} 
    \centering
    \small
    \hspace{-6mm}
\begin{itemize}[leftmargin=0mm]
\setlength{\itemsep}{2pt}
    \item \textbf{Text and Image to Lottie Animation Instruction:} You are a helpful Lottie animation Generator. You output animations in compressed token format using special tokens. You never output JSON format. Generate Lottie animation based on the given image and this description: "\{\textit{caption}\}"
\end{itemize}
\end{tcolorbox}
\vspace{-2mm}
\caption{The instruction for text and image to Lottie animation generation.}
\label{fig:instruction_textimage_to_animation}
\end{minipage}
\vspace{-2mm}
\end{figure}

\begin{figure}[h!]
\centering
\begin{minipage}{0.99\columnwidth}
\vspace{0mm}    
\centering
\begin{tcolorbox} 
    \centering
    \small
    \hspace{-6mm}
\begin{itemize}[leftmargin=0mm]
\setlength{\itemsep}{2pt}
    \item \textbf{Text and Video to Lottie Animation Instruction:} You are a helpful Lottie animation Generator. You output animations in compressed token format using special tokens. You never output JSON format. Generate Lottie animation based on the given video and this description: "\{\textit{caption}\}"
\end{itemize}
\end{tcolorbox}
\vspace{-2mm}
\caption{The instruction for text and video to Lottie animation generation.}
\label{fig:instruction_textvideo_to_animation}
\end{minipage}
\vspace{-4mm}
\end{figure}

\section{User Study}

To validate that LottieBench aligns with human preferences, we conduct a user study following OmniSVG's protocol~\cite{yang2025omnisvg}. We recruit 20 participants with design backgrounds to evaluate outputs from LottieGPT and baselines on three aspects: overall preference, visual vividity, and alignment with input prompts. Participants rate shuffled outputs (without method labels) on a 5-point Likert scale. Tab.~\ref{tab:user_study} shows that LottieGPT achieves the highest scores across all metrics for both static graphics and animation tasks, demonstrating that LottieBench metrics correlate strongly with human judgment.

\begin{table}[htbp]
\centering
\caption{User study results (5-point scale, 20 participants). LottieGPT achieves the highest ratings across all metrics.}
\label{tab:user_study}
\resizebox{\columnwidth}{!}{
\begin{tabular}{l|ccc|ccc}
\hline
\multirow{2}{*}{\textbf{Method}} & \multicolumn{3}{c|}{\textbf{Static Graphics}} & \multicolumn{3}{c}{\textbf{Animation}} \\
& Pref. & Vivid. & Align. & Pref. & Vivid. & Align. \\
\hline
OmniSVG~\cite{yang2025omnisvg} & 3.6 & 3.8 & 3.7 & - & - & - \\
StarVector~\cite{rodriguez2025starvector} & 3.4 & 3.5 & 3.3 & - & - & - \\
\hline
Sora2~\cite{openai2024sora2} & - & - & - & 3.8 & 3.3 & 3.6 \\
Veo 3.1~\cite{google2024veo31} & - & - & - & 3.9 & 3.5 & 3.5 \\
Kling~\cite{kling} & - & - & - & 3.4 & 3.8 & 3.5 \\
\hline
\textbf{LottieGPT} & \textbf{4.1} & \textbf{4.2} & \textbf{4.0} & \textbf{3.8} & \textbf{3.9} & \textbf{3.7} \\
\hline
\end{tabular}
}
\end{table}

\section{Keyframe Easing Interpolation}
\label{sec:keyframe_easing}
Lottie achieves smooth animations through keyframe interpolation with cubic Bézier easing curves. This section explains how time progress is transformed into animation progress using concrete examples from our dataset.

\subsection{Core Concept: Time vs. Animation Progress}
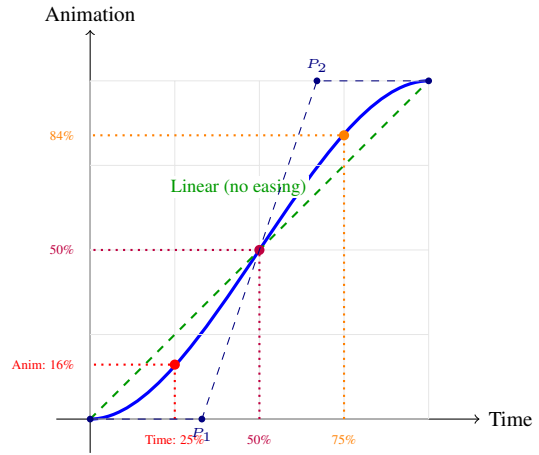
\begin{figure}[h]
\centering
\begin{tikzpicture}[scale=4.5]
    \draw[blue, very thick] (0,0) .. controls (0.33,0) and (0.67,1) .. (1,1);
    
    \draw[->] (-0.1,0) -- (1.15,0) node[right] {\footnotesize Time};
    \draw[->] (0,-0.1) -- (0,1.15) node[above] {\footnotesize Animation};
    
    \draw[gray!20, very thin] (0,0) grid[step=0.25] (1,1);
    
    \draw[green!60!black, dashed, thick] (0,0) -- (1,1) 
        node[pos=0.65, above left, font=\scriptsize, fill=white, inner sep=1pt] {Linear (no easing)};
    
    \draw[red, thick, dotted] (0.25,0) -- (0.25,0.161) -- (0,0.161);
    \fill[red] (0.25,0.161) circle (0.015);
    \node[red, font=\tiny, below] at (0.25,-0.02) {Time: 25\%};
    \node[red, font=\tiny, left] at (-0.02,0.161) {Anim: 16\%};
    
    \draw[purple, thick, dotted] (0.5,0) -- (0.5,0.5) -- (0,0.5);
    \fill[purple] (0.5,0.5) circle (0.015);
    \node[purple, font=\tiny, below] at (0.5,-0.02) {50\%};
    \node[purple, font=\tiny, left] at (-0.02,0.5) {50\%};
    
    \draw[orange, thick, dotted] (0.75,0) -- (0.75,0.839) -- (0,0.839);
    \fill[orange] (0.75,0.839) circle (0.015);
    \node[orange, font=\tiny, below] at (0.75,-0.02) {75\%};
    \node[orange, font=\tiny, left] at (-0.02,0.839) {84\%};
    
    \fill[blue!50!black] (0,0) circle (0.01);
    \fill[blue!50!black] (0.33,0) circle (0.01) node[below, font=\tiny] {$P_1$};
    \fill[blue!50!black] (0.67,1) circle (0.01) node[above, font=\tiny] {$P_2$};
    \fill[blue!50!black] (1,1) circle (0.01);
    \draw[blue!50!black, dashed, thin] (0,0) -- (0.33,0) -- (0.67,1) -- (1,1);
\end{tikzpicture}
\caption{Bézier easing curve transforms time progress (x-axis) into animation progress (y-axis). At 25\% time, animation has only progressed 16\% (slow start); at 75\% time, animation has reached 84\% (slow end).}
\label{fig:time_vs_animation}
\end{figure}

The key to understanding easing is distinguishing between two types of progress:
\textbf{(1) Time Progress} ($t_{\text{norm}}$): Linear progression of time from 0 to 1.
\textbf{(2) Animation Progress} ($t_{\text{eased}}$): Non-linear progression of the animated value from 0 to 1.
The Bézier easing curve maps time progress to animation progress, creating natural-looking motion. Fig.~\ref{fig:time_vs_animation} illustrates this relationship.

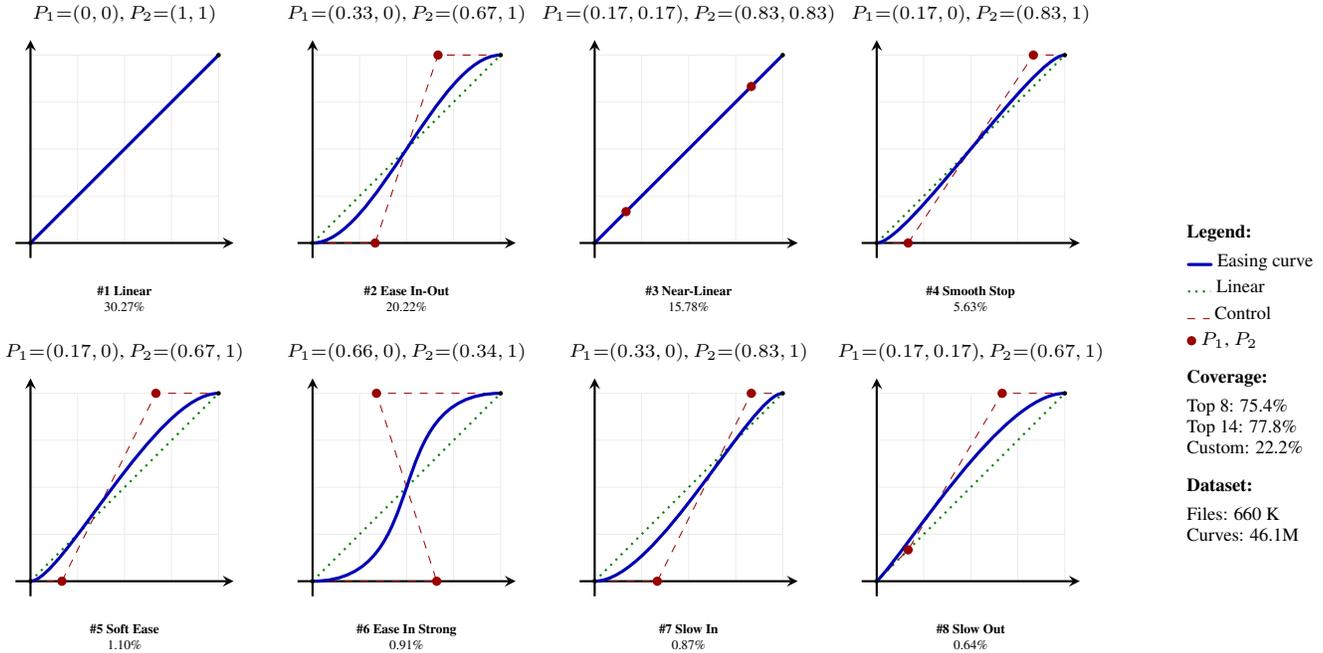
\begin{figure*}[htbp]
\centering
\begin{tikzpicture}[scale=2.5]
    \tikzstyle{curve}=[blue!70!black, very thick, line cap=round]
    \tikzstyle{control}=[red!60!black, dashed, very thin]
    \tikzstyle{grid}=[gray!15, very thin]
    \tikzstyle{axis}=[->, >=stealth, thick]
    \tikzstyle{point}=[circle, fill=red!60!black, inner sep=0pt, minimum size=3.5pt]
    \tikzstyle{label}=[font=\tiny, align=center]
    
    \begin{scope}[shift={(0,0)}]
        \draw[grid] (0,0) grid[step=0.25] (1,1);
        \draw[axis] (-0.08,0) -- (1.08,0);
        \draw[axis] (0,-0.08) -- (0,1.08);
        \draw[green!50!black, dotted, thick] (0,0) -- (1,1);
        \draw[curve] (0,0) -- (1,1);
        \fill[black] (0,0) circle (0.012);
        \fill[black] (1,1) circle (0.012);
        \node[label, below] at (0.5,-0.18) {\textbf{\#1 Linear}\\30.27\%};
        \node[label, above, font=\scriptsize] at (0.5,1.12) {$P_1{=}(0,0)$, $P_2{=}(1,1)$};
    \end{scope}
    
    \begin{scope}[shift={(1.5,0)}]
        \draw[grid] (0,0) grid[step=0.25] (1,1);
        \draw[axis] (-0.08,0) -- (1.08,0);
        \draw[axis] (0,-0.08) -- (0,1.08);
        \draw[green!50!black, dotted, thick] (0,0) -- (1,1);
        \draw[control] (0,0) -- (0.333,0) -- (0.667,1) -- (1,1);
        \draw[curve] (0,0) .. controls (0.333,0) and (0.667,1) .. (1,1);
        \node[point] at (0.333,0) {};
        \node[point] at (0.667,1) {};
        \fill[black] (0,0) circle (0.012);
        \fill[black] (1,1) circle (0.012);
        \node[label, below] at (0.5,-0.18) {\textbf{\#2 Ease In-Out}\\20.22\%};
        \node[label, above, font=\scriptsize] at (0.5,1.12) {$P_1{=}(0.33,0)$, $P_2{=}(0.67,1)$};
    \end{scope}
    
    \begin{scope}[shift={(3,0)}]
        \draw[grid] (0,0) grid[step=0.25] (1,1);
        \draw[axis] (-0.08,0) -- (1.08,0);
        \draw[axis] (0,-0.08) -- (0,1.08);
        \draw[green!50!black, dotted, thick] (0,0) -- (1,1);
        \draw[control] (0,0) -- (0.167,0.167) -- (0.833,0.833) -- (1,1);
        \draw[curve] (0,0) .. controls (0.167,0.167) and (0.833,0.833) .. (1,1);
        \node[point] at (0.167,0.167) {};
        \node[point] at (0.833,0.833) {};
        \fill[black] (0,0) circle (0.012);
        \fill[black] (1,1) circle (0.012);
        \node[label, below] at (0.5,-0.18) {\textbf{\#3 Near-Linear}\\15.78\%};
        \node[label, above, font=\scriptsize] at (0.5,1.12) {$P_1{=}(0.17,0.17)$, $P_2{=}(0.83,0.83)$};
    \end{scope}
    
    \begin{scope}[shift={(4.5,0)}]
        \draw[grid] (0,0) grid[step=0.25] (1,1);
        \draw[axis] (-0.08,0) -- (1.08,0);
        \draw[axis] (0,-0.08) -- (0,1.08);
        \draw[green!50!black, dotted, thick] (0,0) -- (1,1);
        \draw[control] (0,0) -- (0.167,0) -- (0.833,1) -- (1,1);
        \draw[curve] (0,0) .. controls (0.167,0) and (0.833,1) .. (1,1);
        \node[point] at (0.167,0) {};
        \node[point] at (0.833,1) {};
        \fill[black] (0,0) circle (0.012);
        \fill[black] (1,1) circle (0.012);
        \node[label, below] at (0.5,-0.18) {\textbf{\#4 Smooth Stop}\\5.63\%};
        \node[label, above, font=\scriptsize] at (0.5,1.12) {$P_1{=}(0.17,0)$, $P_2{=}(0.83,1)$};
    \end{scope}
    
    \begin{scope}[shift={(0,-1.8)}]
        \draw[grid] (0,0) grid[step=0.25] (1,1);
        \draw[axis] (-0.08,0) -- (1.08,0);
        \draw[axis] (0,-0.08) -- (0,1.08);
        \draw[green!50!black, dotted, thick] (0,0) -- (1,1);
        \draw[control] (0,0) -- (0.167,0) -- (0.667,1) -- (1,1);
        \draw[curve] (0,0) .. controls (0.167,0) and (0.667,1) .. (1,1);
        \node[point] at (0.167,0) {};
        \node[point] at (0.667,1) {};
        \fill[black] (0,0) circle (0.012);
        \fill[black] (1,1) circle (0.012);
        \node[label, below] at (0.5,-0.18) {\textbf{\#5 Soft Ease}\\1.10\%};
        \node[label, above, font=\scriptsize] at (0.5,1.12) {$P_1{=}(0.17,0)$, $P_2{=}(0.67,1)$};
    \end{scope}
    
    \begin{scope}[shift={(1.5,-1.8)}]
        \draw[grid] (0,0) grid[step=0.25] (1,1);
        \draw[axis] (-0.08,0) -- (1.08,0);
        \draw[axis] (0,-0.08) -- (0,1.08);
        \draw[green!50!black, dotted, thick] (0,0) -- (1,1);
        \draw[control] (0,0) -- (0.660,0) -- (0.340,1) -- (1,1);
        \draw[curve] (0,0) .. controls (0.660,0) and (0.340,1) .. (1,1);
        \node[point] at (0.660,0) {};
        \node[point] at (0.340,1) {};
        \fill[black] (0,0) circle (0.012);
        \fill[black] (1,1) circle (0.012);
        \node[label, below] at (0.5,-0.18) {\textbf{\#6 Ease In Strong}\\0.91\%};
        \node[label, above, font=\scriptsize] at (0.5,1.12) {$P_1{=}(0.66,0)$, $P_2{=}(0.34,1)$};
    \end{scope}
    
    \begin{scope}[shift={(3,-1.8)}]
        \draw[grid] (0,0) grid[step=0.25] (1,1);
        \draw[axis] (-0.08,0) -- (1.08,0);
        \draw[axis] (0,-0.08) -- (0,1.08);
        \draw[green!50!black, dotted, thick] (0,0) -- (1,1);
        \draw[control] (0,0) -- (0.333,0) -- (0.833,1) -- (1,1);
        \draw[curve] (0,0) .. controls (0.333,0) and (0.833,1) .. (1,1);
        \node[point] at (0.333,0) {};
        \node[point] at (0.833,1) {};
        \fill[black] (0,0) circle (0.012);
        \fill[black] (1,1) circle (0.012);
        \node[label, below] at (0.5,-0.18) {\textbf{\#7 Slow In}\\0.87\%};
        \node[label, above, font=\scriptsize] at (0.5,1.12) {$P_1{=}(0.33,0)$, $P_2{=}(0.83,1)$};
    \end{scope}
    
    \begin{scope}[shift={(4.5,-1.8)}]
        \draw[grid] (0,0) grid[step=0.25] (1,1);
        \draw[axis] (-0.08,0) -- (1.08,0);
        \draw[axis] (0,-0.08) -- (0,1.08);
        \draw[green!50!black, dotted, thick] (0,0) -- (1,1);
        \draw[control] (0,0) -- (0.167,0.167) -- (0.667,1) -- (1,1);
        \draw[curve] (0,0) .. controls (0.167,0.167) and (0.667,1) .. (1,1);
        \node[point] at (0.167,0.167) {};
        \node[point] at (0.667,1) {};
        \fill[black] (0,0) circle (0.012);
        \fill[black] (1,1) circle (0.012);
        \node[label, below] at (0.5,-0.18) {\textbf{\#8 Slow Out}\\0.64\%};
        \node[label, above, font=\scriptsize] at (0.5,1.12) {$P_1{=}(0.17,0.17)$, $P_2{=}(0.67,1)$};
    \end{scope}
    
    \begin{scope}[shift={(6.1,-0.8)}]
        \node[label, anchor=west, align=left, font=\scriptsize] at (0,0) {
            \textbf{Legend:}\\[3pt]
            \tikz{\draw[curve] (0,0) -- (0.3,0);} Easing curve\\[2pt]
            \tikz{\draw[green!50!black, dotted, thick] (0,0) -- (0.3,0);} Linear\\[2pt]
            \tikz{\draw[control] (0,0) -- (0.3,0);} Control\\[2pt]
            \tikz{\node[point] at (0.15,0) {};} $P_1$, $P_2$\\[6pt]
            \textbf{Coverage:}\\[3pt]
            Top 8: 75.4\%\\
            Top 14: 77.8\%\\
            Custom: 22.2\%\\[6pt]
            \textbf{Dataset:}\\[3pt]
            Files: 660 K\\
            Curves: 46.1M\\
        };
    \end{scope}
\end{tikzpicture}
\caption{Top 8 most common Bézier easing curves in Lottie animations, covering 75.4\% of all usage. Each curve is defined by control points $P_1=(o_x, o_y)$ and $P_2=(i_x, i_y)$, with fixed endpoints at $(0,0)$ and $(1,1)$. Green dotted line shows linear interpolation for comparison. The legend on the right provides dataset statistics and visual element descriptions. The third most common pattern (15.78\%) uses control points $(0.167, 0.167)$ and $(0.833, 0.833)$, which lie on the linear diagonal, making it functionally identical to \#1 Linear. This redundancy represents a significant compression opportunity.}
\label{fig:easing_top8}
\end{figure*}

\subsection{Bézier Curve Definition}
The easing function maps normalized time progress to animation progress:
\begin{equation}
t_{\text{eased}} = f(t_{\text{norm}}), \quad t_{\text{norm}}, t_{\text{eased}} \in [0,1]
\end{equation}
This function is defined implicitly via a cubic Bézier curve with fixed endpoints $P_0=(0,0)$, $P_3=(1,1)$ and control points $P_1=(o_x, o_y)$, $P_2=(i_x, i_y)$:
\begin{equation}
\begin{split}
\mathbf{B}(u) = &(1-u)^3 P_0 + 3(1-u)^2 u P_1 \\
                &+ 3(1-u) u^2 P_2 + u^3 P_3, \quad u \in [0,1]
\end{split}
\end{equation}
Expanding with $P_0=(0,0)$ and $P_3=(1,1)$:
\begin{align}
t_{\text{norm}} &= x(u) = 3(1-u)^2 u \cdot o_x + 3(1-u) u^2 \cdot i_x + u^3 \label{eq:bezier_x} \\
t_{\text{eased}} &= y(u) = 3(1-u)^2 u \cdot o_y + 3(1-u) u^2 \cdot i_y + u^3 \label{eq:bezier_y}
\end{align}
To evaluate $f(t_{\text{norm}})$: solve Eq.~\eqref{eq:bezier_x} for $u$, then compute Eq.~\eqref{eq:bezier_y}. This is necessary because the Bézier curve provides a parametric (not explicit) definition of $f$.

\subsection{Top 8 Easing Curves in Lottie Animations}

Fig.~\ref{fig:easing_top8} shows the eight most common easing curves from our analysis of 46.1M curves across 660K animation files. These patterns account for 75.4\% of all usage, revealing clear preferences and encoding redundancies: Linear (30.27\%) uses identity control points, while the third pattern (15.78\%) is mathematically equivalent to linear despite explicit Bézier parameters $(0.167, 0.167)$ and $(0.833, 0.833)$. Together, these represent 46.05\% linear motion.
The second pattern (20.22\%) provides standard ease-in-ease-out motion with control points $(0.333, 0)$ and $(0.667, 1)$, mimicking real-world physics.

This high repetition rate presents a significant compression opportunity. The top 14 presets cover 77.8\% of all curves, suggesting that a token-based encoding scheme could substantially improve storage density by replacing redundant Bézier parameters with compact preset identifiers.

\begin{figure*}[htbp]
\centering
\begin{minipage}[t]{0.235\textwidth}
    \centering
    \includegraphics[width=\linewidth]{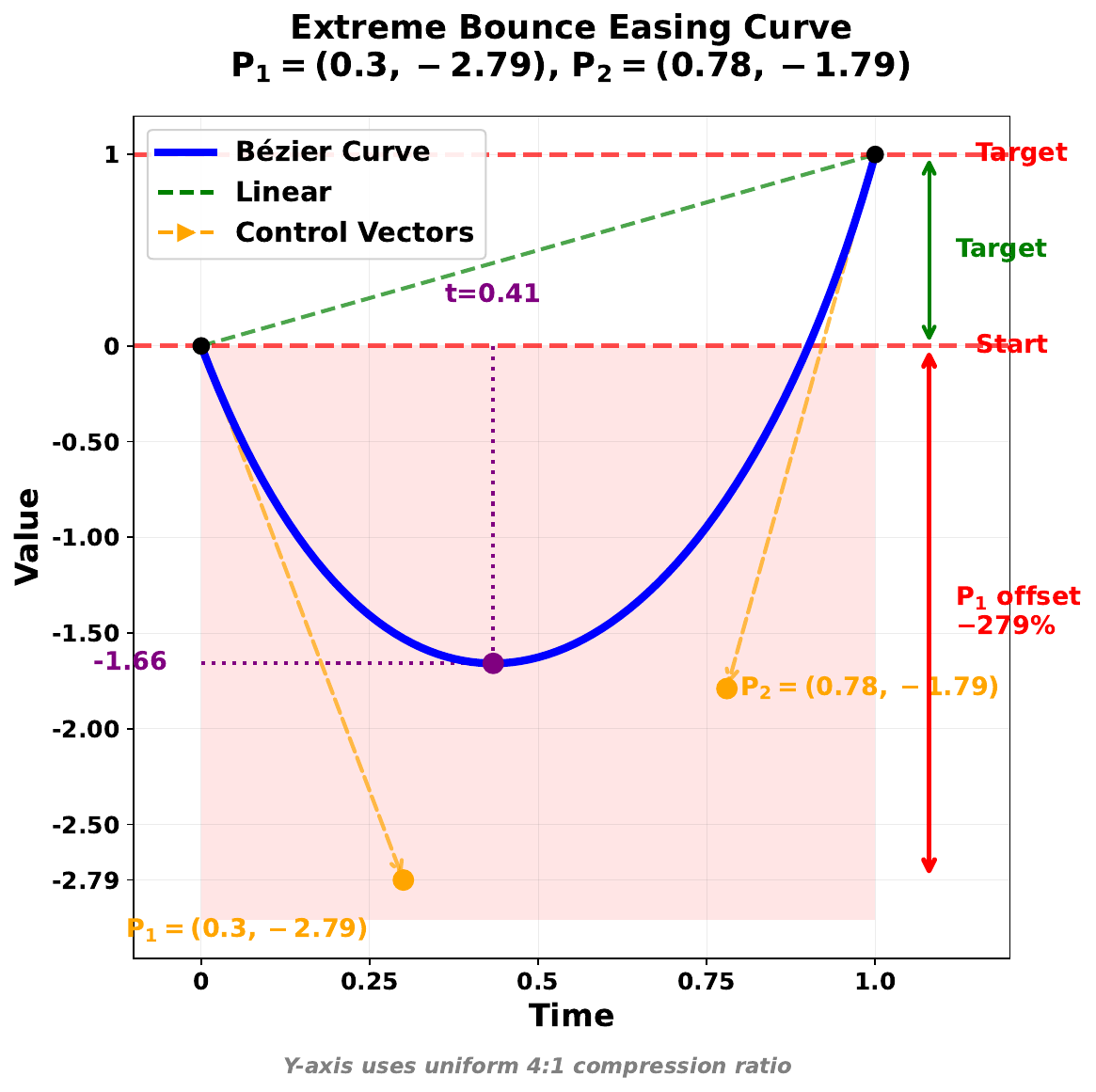}
    \subcaption{Bézier curve with extreme undershoot}
    \label{fig:extreme_bounce_curve}
\end{minipage}%
\hfill
\begin{minipage}[t]{0.235\textwidth}
    \centering
    \includegraphics[height=\linewidth]{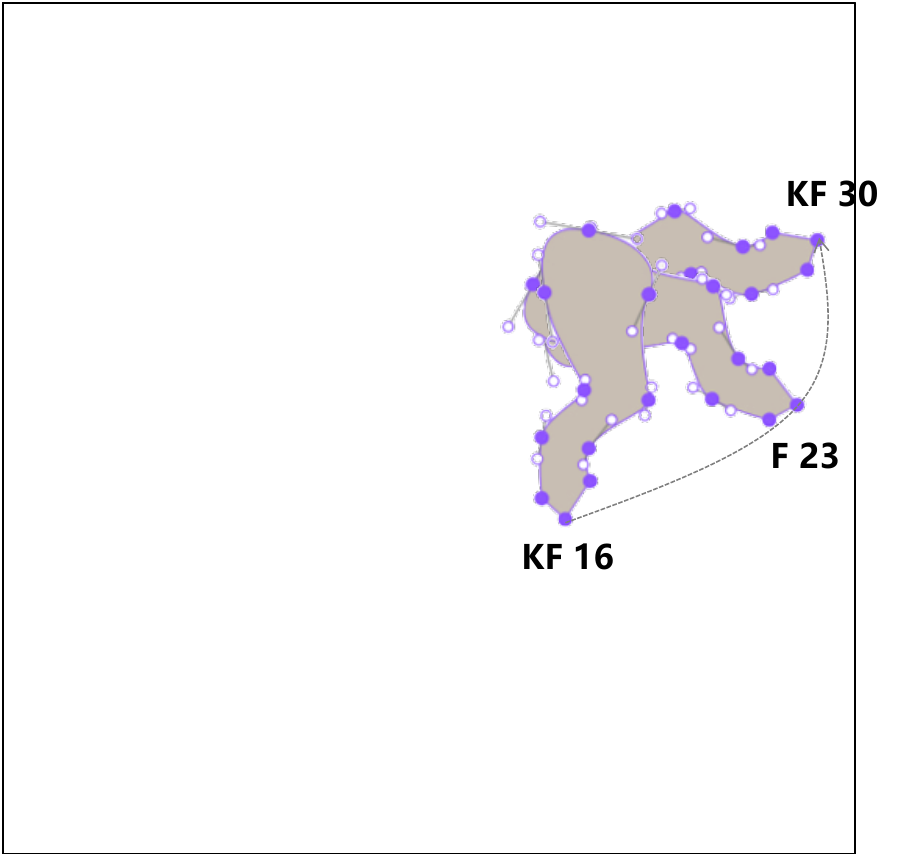}
    \subcaption{Keyframes 16 and 30, interpolated frame 23}
    \label{fig:bounce_easing_1}
\end{minipage}%
\hfill
\begin{minipage}[t]{0.235\textwidth}
    \centering
    \includegraphics[height=\linewidth]{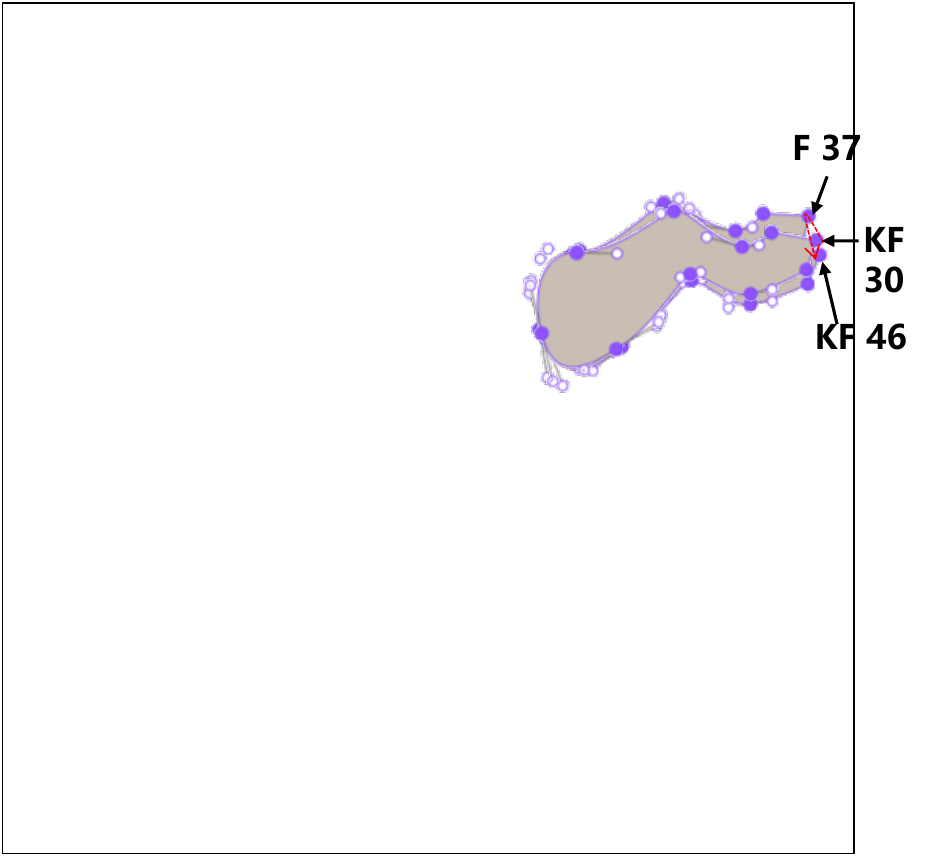}
    \subcaption{Keyframes 30 and 46, interpolated frame 37 (bounce)}
    \label{fig:bounce_easing_2}
\end{minipage}%
\hfill
\begin{minipage}[t]{0.235\textwidth}
    \centering
    \includegraphics[height=\linewidth]{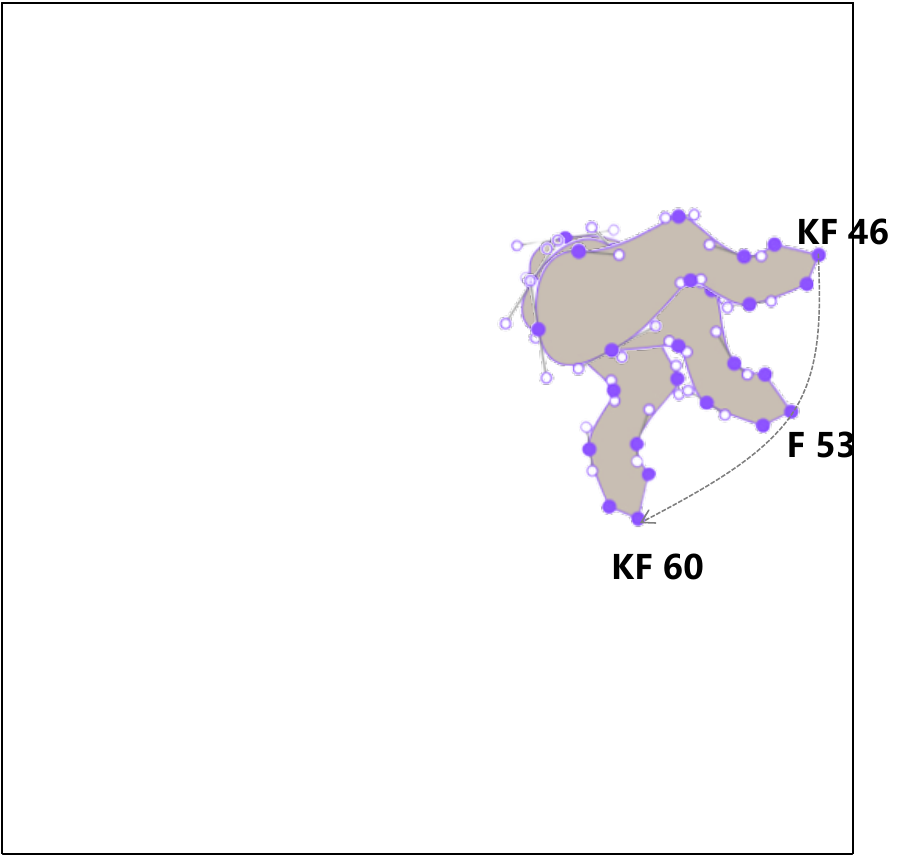}
    \subcaption{Keyframes 46 and 60, interpolated frame 53}
    \label{fig:bounce_easing_3}
\end{minipage}

\caption{Bounce easing animation demonstrating extreme undershoot. (a) Bézier curve with $P_1=(0.3, -2.79)$ and $P_2=(0.78, -1.79)$ reaching minimum $y=-1.6583$ at $t_{\text{norm}}=0.4134$ (Y-axis compressed 4:1 for visibility). (b) Normal interpolation between keyframes 16 and 30. (c) Bounce segment between keyframes 30 and 46, showing extreme undershoot at frame 37 where the $3.9°$ target rotation is amplified to a $6.47°$ reverse motion (165.8\% overshoot, actual minimum at frame 36.61). (d) Settling segment between keyframes 46 and 60. We removed the layer-level animations and shape-level position animations from the original animation, retaining only the shape-level rotation animations.}
\label{fig:bounce_easing_combined}
\end{figure*}

\subsection{Bounce Easing Example}
\label{subsec:bounce_easing}
Lottie's Bézier easing curves support control points with $y$-coordinates outside $[0,1]$, enabling advanced effects such as bounce and overshoot animations. This section demonstrates how such curves are interpolated to create dramatic spring-like motion.
We demonstrate the easing interpolation function using the example from Fig.~\ref{fig:lottie_tokenizer}, focusing on the interval between frame 30 and frame 46. Fig.~\ref{fig:bounce_easing_combined} illustrates the Bézier curve of this spring motion effect in detail.
Consider a rotation animation from $-113.4°$ to $-109.5°$ over frames 30-46 (16 frames) with extreme bounce easing defined by control points $P_1=(0.3, -2.79)$ and $P_2=(0.78, -1.79)$:
\begin{lstlisting}[style=json,basicstyle=\small\ttfamily]
"r": {
  "a": 1,
  "k": [
    {"t": 30, "s": [-113.4], 
     "o": {"x": [0.3], "y": [-2.79]}, 
     "i": {"x": [0.78], "y": [-1.79]}},
    {"t": 46, "s": [-109.5],
     "o": {"x": [0.35], "y": [0.16]},
     "i": {"x": [0.83], "y": [0.87]}}
  ]
}
\end{lstlisting}
Note that the first keyframe uses extreme negative $y$-values in its outgoing control points, creating the bounce effect, while the second keyframe uses standard positive values for the next segment.

\noindent\paragraph{Interpolation Process.}
To compute the rotation at frame $f = 37$, we follow a four-step process:

\noindent\textbf{Step 1: Normalize time progress}
\begin{equation}
t_{\text{norm}} = \frac{f - t_{\text{start}}}{t_{\text{end}} - t_{\text{start}}} = \frac{37 - 30}{46 - 30} = \frac{7}{16} = 0.4375
\end{equation}

\noindent\textbf{Step 2: Solve for curve parameter $u$}
Find $u$ such that $x(u) = t_{\text{norm}}$ using Eq.~\eqref{eq:bezier_x}. For $t_{\text{norm}} = 0.4375$:
\begin{equation}
3(1-u)^2 u \cdot 0.3 + 3(1-u) u^2 \cdot 0.78 + u^3 = 0.4375
\end{equation}
Using Newton-Raphson iteration yields $u \approx 0.4603$.

\noindent\textbf{Step 3: Evaluate animation progress}
Compute $t_{\text{eased}} = y(u)$ using Eq.~\eqref{eq:bezier_y} with the negative control point $y$-values:
\begin{equation}
\begin{split}
t_{\text{eased}} &= 3(1-0.4603)^2 \cdot 0.4603 \cdot (-2.79) \\
                 &\quad + 3(1-0.4603) \cdot (0.4603)^2 \cdot (-1.79) + (0.4603)^3 \\
&\approx -1.6531
\end{split}
\end{equation}
The negative value indicates the animation has reversed by 165.31\% of its total range.

\noindent\textbf{Step 4: Interpolate final value}
The rotation range is $\Delta r = -109.5 - (-113.4) = 3.9°$:
\begin{equation}
\begin{split}
\text{rotation}(37) &= s_{\text{start}} + (s_{\text{end}} - s_{\text{start}}) \times t_{\text{eased}} \\
&= -113.4 + 3.9 \times (-1.6531) \\
&= -113.4 - 6.45 \\
&= -119.85°
\end{split}
\end{equation}
Despite the animation's target being a small $3.9°$ clockwise rotation (from $-113.4°$ to $-109.5°$), the bounce effect causes the object to first rotate $6.45°$ counterclockwise to $-119.85°$ before rebounding to the final position. The actual minimum occurs at frame 36.61 ($t_{\text{norm}} = 0.4134$) with $t_{\text{eased}} = -1.6583$ and rotation $-119.87°$.

\begin{table}[h]
\centering
\caption{Bounce animation progression with extreme undershoot easing}
\label{tab:bounce_progress}
\resizebox{\columnwidth}{!}{%
\begin{tabular}{@{}ccccp{3.2cm}@{}}
\toprule
\textbf{Frame} & $\boldsymbol{t_{\text{norm}}}$ & $\boldsymbol{t_{\text{eased}}}$ & \textbf{Rotation} & \textbf{Phase} \\
\midrule
30    & 0.0000 & 0.0000   & $-113.40°$ & Initial state \\
35    & 0.3125 & $-1.5663$ & $-119.51°$ & $\rotatebox[origin=c]{-90}{$\curvearrowleft$}$ \\
36    & 0.3750 & $-1.6453$ & $-119.82°$ & $\rotatebox[origin=c]{-90}{$\curvearrowleft$}$ \\
36.61 & 0.4134 & $-1.6583$ & $-119.87°$ & $\rotatebox[origin=c]{-90}{$\curvearrowleft$}$  \\
37    & 0.4375 & $-1.6531$ & $-119.85°$ & Velocity $\approx 0$ \\
38    & 0.5000 & $-1.5925$ & $-119.61°$ & $\rotatebox[origin=c]{-90}{$\curvearrowright$}$ \\
40    & 0.6250 & $-1.2781$ & $-118.38°$ & $\rotatebox[origin=c]{-90}{$\curvearrowright$}$ \\
43.6  & 0.8500 & $-0.1279$ & $-113.90°$ & $\rotatebox[origin=c]{-90}{$\curvearrowright$}$ \\
46    & 1.0000 & 1.0000   & $-109.50°$ & Settle at target \\
\bottomrule
\end{tabular}%
}
\end{table}

\noindent\paragraph{Animation Phases.}
Tab.~\ref{tab:bounce_progress} shows the complete bounce trajectory through distinct phases:
\begin{itemize}
    \item \textbf{Undershoot phase} (frames 30-36.61): Despite a target of only $3.9°$ clockwise rotation, the object rotates $6.47°$ counterclockwise (165.8\% overshoot), reaching maximum at frame 36.61 where $t_{\text{eased}} = -1.6583$
    \item \textbf{Velocity zero point}: At frame 37 ($t_{\text{norm}} = 0.4375$, $u \approx 0.4603$), the curve's derivative $\frac{dy}{du} \approx 0$, marking the transition from CCW to CW motion. This occurs slightly after the minimum point.
    \item \textbf{Rebound phase} (frames 37-43.6): Clockwise acceleration crosses the starting position at frame 43.6
    \item \textbf{Settling phase} (frames 43.6-46): Deceleration to final position at $-109.5°$
\end{itemize}

This spring-like behavior is achieved through negative $y$-values in control points, creating animation progress values outside $[0,1]$. The extreme undershoot of 165.8\% demonstrates how Bézier easing can amplify small rotations into dramatic bounce effects, commonly used for impact animations, dramatic transitions, and physics-based UI feedback.

\section{Limitations and Future Work}

While Lottie Animation can compress After Effects animations into a compact representation and achieve higher compression ratios compared to SVG, and our proposed Lottie Tokenizer has been validated on existing VLMs to generate Lottie animations from text and image inputs, there remain several key limitations that warrant future investigation.

\textbf{Limited Color Representation.}
A fundamental constraint of all vector graphics and vector animations (including SVG, Lottie, and HTML/CSS animations) is their inability to express the full range of colors present in real-world imagery. Vector graphics typically represent colors by filling shapes with solid colors or regular gradients, which limits their capacity to capture complex color gradients, textures, and photorealistic details. This inherent limitation makes vector-based representations less suitable for tasks requiring high-fidelity color reproduction or photorealistic rendering. Future work could explore hybrid representations that combine the compactness of vector formats with richer color modeling capabilities, such as incorporating procedural texture representations.

\textbf{Complex Animation Effects.}
While the current Lottie format is powerful for many animation scenarios, it has limitations in representing certain advanced effects commonly used in professional animation workflows, such as complex particle systems, intricate shape paths, advanced blending modes, and 3D transformations. The existing Lottie tokenizer still has low compression efficiency for these components, and future work will need to extend the Lottie tokenizer accordingly.

\textbf{Temporal Coherence and Long Animations.}
Although our model can generate animations with reasonable temporal coherence, generating very long and complex animations remains challenging due to the context length limitations of VLM generation.
Future work could explore hierarchical generation strategies to better model long-range dependencies in animation sequences.

\textbf{Future Work.}
Despite these limitations, After Effects animations and the Lottie format continue to find widespread applications across numerous domains, including UI/UX design, web frontend development, 2D animated films, motion graphics, and interactive media. The compact representation, editability, and scalability of Lottie animations make them particularly valuable for these applications. Moving forward, we aim to extend LottieGPT to express more complex and realistic animations while maintaining the advantages of vector-based representations. We believe that continued advancement in vector animation generation will unlock new possibilities for creative content creation and interactive media applications.

\end{document}